\RequirePackage{fix-cm}
\documentclass[twocolumn]{svjour3}          

\smartqed  
\usepackage{graphicx}
\usepackage{amsmath,amssymb} 
\usepackage[utf8x]{inputenc}
\usepackage{pgfplots}
\pgfplotsset{compat=newest}
\usepackage{tikz}
\usetikzlibrary{backgrounds}
\usepackage{color}
\usepackage{xcolor}
\usepackage{booktabs,siunitx} 
\usepackage{makecell}
\usepackage[caption=false]{subfig}
\usepackage[noadjust]{cite}
\usepackage[colorlinks=true,allcolors=green]{hyperref}
\usepackage{rotating}

\usepackage{times}
\usepackage{pifont}
\usepackage{arydshln}
\usepackage{adjustbox}
\usepackage{array,multirow}

\usepackage{epsfig}
\usepackage{paralist}
\usepackage[english]{babel}

\usepackage{authblk}
\usepackage{diagbox}

\usepackage{natbib}
\setcitestyle{authoryear,open={(},close={)}}

\begin{document}

\title{PointSea: Point Cloud Completion via Self-structure Augmentation}

\titlerunning{PointSea: Point Cloud Completion via Self-structure Augmentation}

\author{Zhe Zhu   \and
        Honghua Chen \and
        Xing He \and
        Mingqiang Wei
}

\institute{              
                Zhe Zhu \at zhuzhe0619@nuaa.edu.cn \\ \\
                Honghua Chen$^{\dag}$ \at chenhonghuacn@gmail.com \\ \\
                Xing He \at hexing@nuaa.edu.cn \\ \\
                Mingqiang Wei$^{\dag}$ \at mqwei@nuaa.edu.cn \\ \\
              Nanjing University of Aeronautics and Astronautics \\
              Nanjing, China \\ \\
              $^{\dag}$ Corresponding authors
%
}
\date{Received: date / Accepted: date}

\maketitle

\begin{abstract}
\sloppy{  
Point cloud completion is a fundamental yet not well-solved problem in 3D vision.
Current approaches often rely on 3D coordinate information and/or additional data (e.g., images and scanning viewpoints) to fill in missing parts.  
Unlike these methods, we explore self-structure augmentation and propose \textbf{PointSea} for global-to-local point cloud completion. 
In the global stage, consider how we inspect a defective region of a physical object, we may observe it from various perspectives for a better understanding.
Inspired by this, PointSea augments data representation by leveraging self-projected depth images from multiple views. To reconstruct a compact global shape from the cross-modal input, we incorporate a feature fusion module to fuse features at both intra-view and inter-view levels.
In the local stage, to reveal highly detailed structures, we introduce a point generator called the self-structure dual-generator. This generator integrates both learned shape priors and geometric self-similarities for shape refinement. Unlike existing efforts that apply a unified strategy for all points, our dual-path design adapts refinement strategies conditioned on the structural type of each point, addressing the specific incompleteness of each point. 
Comprehensive experiments on widely-used benchmarks demonstrate that PointSea effectively understands global shapes and generates local details from incomplete input, showing clear improvements over existing methods. 
Our code is available at \emph{\textcolor{magenta}{https://github.com/czvvd/SVDFormer\_PointSea}}.
}
\end{abstract}

\section{Introduction}
\begin{figure*}[h]
    \hsize=\textwidth
    \centering
    \includegraphics[width=\textwidth]{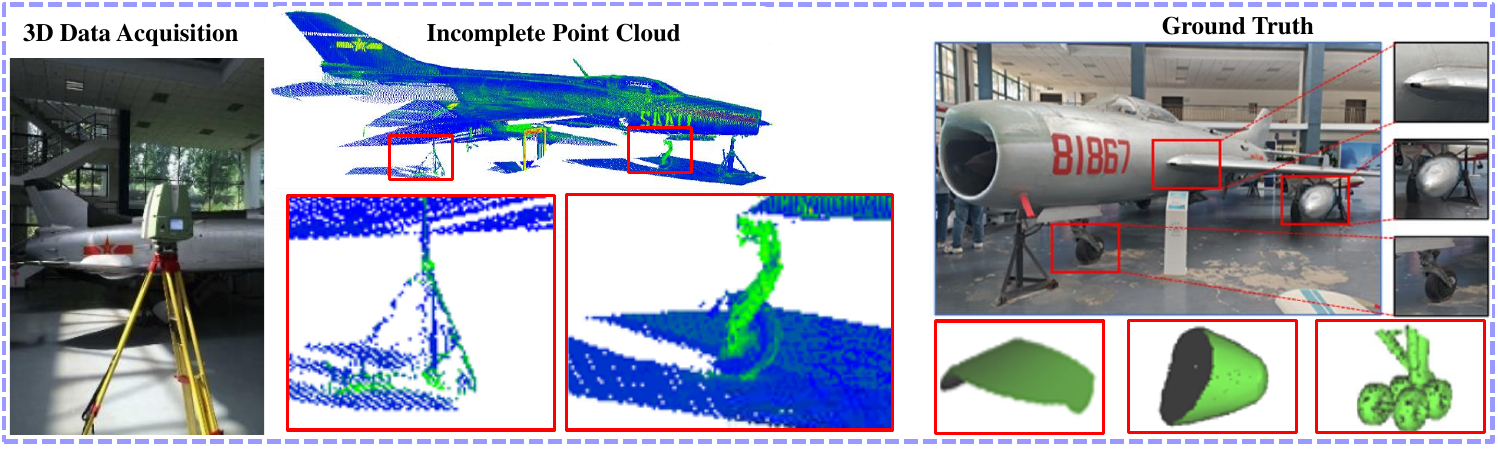}
    \caption{The original point cloud of large and complex objects is often incomplete due to many factors. Point cloud completion plays an essential role in extensive practical applications.}
    \label{fig:incomplete}
\end{figure*}

Raw-captured point clouds are often incomplete due to factors like occlusion, surface reflectivity, and limited scanning range (see Fig. \ref{fig:incomplete}). Before they can be used in downstream applications (e.g., digital twin), they need to be faithfully completed, a process known as point cloud completion. Recent years have witnessed significant progress in this field~\citep{yuan2018pcn,huang2020pf,zhang2020detail,yu2021pointr,9928787,yan2022fbnet,zhang2022shape,tang2022lake,zhou2022seedformer,zhang2022point,10232862,SCRN}. 
However, the sparsity and large structural incompleteness of point clouds still limit their ability to produce satisfactory results. There are two primary challenges in point cloud completion:
\begin{itemize}
    \item Crucial semantic parts are often absent in the partial observations.
    \item Detailed structures cannot be effectively recovered.
\end{itemize}

\begin{figure*}[h]
    \hsize=\textwidth
    \centering
    \includegraphics[width=\textwidth]{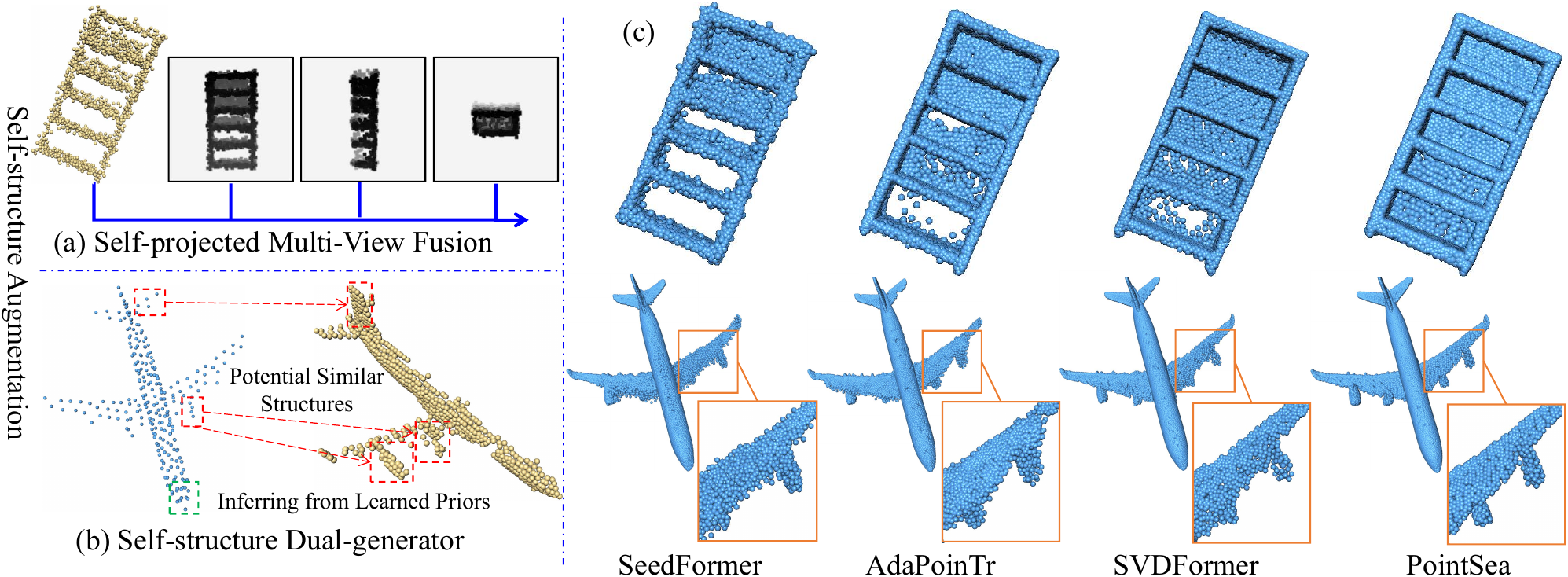}
    \caption{We explore Self-structure Augmentation, a novel strategy for high-quality global-to-local point cloud completion, dubbed PointSea.  (a) In the global stage, PointSea understands incomplete shapes from self-projected multiple views. (b) In the local stage, PointSea collaborates both similar geometric similarities in input (red boxes) and learned shape priors (green boxes) for shape refinement. (c) PointSea shows clear improvements over its competitors, including SeedFormer~\citep{zhou2022seedformer}, AdaPoinTr~\citep{10232862}, and SVDFormer~\citep{Zhu_2023_ICCV}.}
    \label{fig:teaser}
\end{figure*}

The first challenge leads to a vast solution space for point-based networks~\citep{yuan2018pcn,9928787,yu2021pointr,zhou2022seedformer} to robustly locate missing regions and create a partial-to-complete mapping.
Some alternative methods attempt to address this issue by incorporating additional color images~\citep{zhang2021view,aiello2022crossmodal,zhu2023csdn} or viewpoints~\citep{zhang2022shape,Gong_2021_ICCV,fu2023vapcnet}. 
However, the paired images with well-calibrated camera parameters are hard to obtain, as well as the scanning viewpoints.
To resolve the second challenge, some recent approaches~\citep{9928787,yan2022fbnet} utilize skip-connections between multiple refinement steps, allowing them to iteratively recover finer details with learned shape pattern priors. Some other methods prioritize preserving the original detail information by no-pooling encoding~\citep{zhou2022seedformer} or structural relational enhancement~\citep{10106495}.
However, all the above approaches typically employ a unified refinement strategy for all surface regions, making it difficult to infer various detailed structures. By observing and analyzing different kinds of partial point clouds, we find that the missing surface regions can be classified into two types.
The first type lacks similar structures in the input shape, and their reconstruction heavily relies on the learned shape prior.
The second type is consistent with the local structures present in the partial input, and their recovery can be facilitated by appropriate geometric regularity~\citep{zhao2021sign}. 
For instance, large-scale LiDAR scans are notably sparse and contain limited information for generating fine details. Indoor depth scans, conversely, capture more semantic information, consequently providing richer geometric cues for completion.

Based on the above observations, we propose a novel neural network for point cloud completion called PointSea.
Our method improves the global-to-local paradigm by fully leveraging self-structure information. 

First, similar to how a human would perceive and locate the missing areas of a physical object by observing it from different viewpoints, we aim to drive the neural network to absorb human knowledge by augmenting the data representation. To achieve it, we design a Self-view Fusion Network (SVFNet) that learns an effective descriptor, well depicting the global shape from both the point cloud data and depth maps captured from multiple viewpoints (see Fig.~\ref{fig:teaser} (a)). To better exploit such kind of cross-modal information, we introduce a feature fusion module to improve the discriminative power of multi-view features.

Regarding the second challenge, our insight is to disentangle refinement strategies conditioned on the structural type of each point to reveal detailed geometric structures.
Therefore, we design a Self-structure Dual-Generator (SDG) with a pair of parallel refinement units, called \emph{Structure Analysis} and \emph{Similarity Alignment}, respectively. 
As depicted in Fig.~\ref{fig:teaser} (b), two units function distinctly during refinement.
The former unit analyzes the generated coarse point clouds by explicitly encoding local incompleteness, enabling it to match learned geometric patterns from training data to infer underlying shapes.
The \emph{Similarity Alignment} unit finds the features of similar structures for every point, thus making it easier to refine its local region by mimicking the input's local structures. 
With this dual-path design, our method can generate reasonable results for various types of input shapes, including symmetrical synthetic models with different degrees of incompleteness and highly sparse real-world scans.

We conduct comprehensive evaluations of PointSea across multiple datasets, including PCN, ShapeNet-55/34, Projected-ShapeNet-55/34, KITTI, ScanNet, and Matterport3D, with a specific focus on object shape completion. Additionally, we extend our analysis to assess the performance of PointSea for semantic scene completion. Experimental results demonstrate that PointSea achieves state-of-the-art performance on these benchmarks.
Our contributions are listed below:
\begin{itemize}
  \item 
  We propose a novel point cloud completion network based on self-structure augmentation, which significantly improves the recovery of both global shapes and local details in point clouds.
  \item 
  We propose a Self-view Fusion Network (SVFNet) to enhance multi-view and cross-modal features, resulting in a more plausible global shape.
  \item 
  We propose a Self-structure Dual-Generator (SDG) to refine the global shape. This component handles various kinds of incomplete shapes by jointly learning local pattern priors and self-similarities.
\end{itemize}

A preliminary version of this work was presented at \textit{ICCV 2023}~\citep{Zhu_2023_ICCV}. We extend the conference version by introducing the following new contributions:

\begin{itemize}
  \item While SVDFormer~\citep{Zhu_2023_ICCV} has already demonstrated commendable performance, some of the design choices are sub-optimal. Therefore, we propose an improved version of SVDFormer, called PointSea. For example, in the global stage, SVDFormer considers feature fusion at only the intra-view level. To achieve a more enriched cross-modal feature representation, PointSea incorporates a feature fusion strategy at both inter-view and intra-view levels. In the local refinement stage, SVDFormer assigns equal weight to two refinement paths and combines their outputs directly. However, as discussed earlier, we acknowledge that the significance of these two paths varies across different missing regions. Therefore, we propose a path selection module to adaptively aggregate diverse features. These new designs in PointSea lead to a significant performance enhancement compared to its predecessor.

 \item We conduct a wider series of experiments, specifically focusing on real-world conditions such as point cloud semantic scene completion and a more systematic evaluation of real-world object scans. Finally, we provide more in-depth empirical analyses and visualizations.
\end{itemize}
\label{secIntroduction}

\section{Related Work}
\subsection{Learning-based Shape Completion}
Early learning-based methods ~\citep{dai2017shape,han2017high,stutz2018learning,varley2017shape} often rely on voxel-based representations for 3D convolutional neural networks. However, these approaches are limited by their high computational cost and limited resolution. Alternatively, GRNet~\citep{xie2020grnet} and VE-PCN~\citep{wang2021voxel} use 3D grids as an intermediate representation for point-based completion.  
In recent years, several methods have been proposed to directly process points by end-to-end networks. One pioneering point-based work is PCN~\citep{yuan2018pcn}, which uses a shared multi-layer perceptron (MLP) to extract features and generates additional points using a folding operation~\citep{yang2018foldingnet} in a coarse-to-fine manner. Inspired by it, a lot of point-based methods~\citep{wang2020cascaded,liu2020morphing,wen2020point,9928787,zhou2022seedformer,yu2021pointr,SCRN,pan2020ecg,AGConv,scpnet,zhou2020geometry,chen2022repcd} have been proposed.

Later, to address the issue of limited information available in partial shapes, several works~\citep{zhang2021view,aiello2022crossmodal,zhu2023csdn,huang2022spovt,zhang2022shape,Gong_2021_ICCV,fu2023vapcnet} have explored the use of auxiliary data to enhance performance. 
Some approaches~\citep{zhang2021view,aiello2022crossmodal,zhu2023csdn} involve the combination of rendered color images and partial point clouds, along with the corresponding camera parameters.
Another line of works~\citep{zhang2022shape,Gong_2021_ICCV,fu2023vapcnet} seeks to utilize the scanning viewpoints. They either produce points along the viewpoints~\citep{Gong_2021_ICCV,zhang2022shape} or use the viewpoints for self-supervised pretraining~\citep{fu2023vapcnet}.
Semantic labels are also used in a recent approach~\citep{huang2022spovt} to provide strong priors.
Although these methods have shown promising results, they often require additional input that is difficult to obtain in practical settings. 
Different from these 3D data-driven methods, MVCN~\citep{hu2019render4completion} operates completion solely in the 2D domain using a conditional GAN. However, it cannot supervise the results using ground truth with rich space information. 
In contrast to these methods, we propose to integrate the representation abilities of 3D and 2D modalities by observing self-structures. As a result, our method achieves a more comprehensive perception of the overall shape without requiring additional information.

Considering the high-quality details generation, a variety of strategies have been introduced by learning shape context and local spatial relationships. 
To achieve this goal, state-of-the-art methods design various refinement modules to learn better shape priors from the training data.
SnowflakeNet~\citep{9928787} introduces Snowflake Point Deconvolution (SPD), which leverages skip-transformer to model the relation between parent points and child points. 
FB-Net~\citep{yan2022fbnet} adopts the feedback mechanism during refinement and generates points recurrently.
LAKe-Net~\citep{tang2022lake} integrates its surface-skeleton representation into the refinement stage, which makes it easier to learn the missing topology part.
Another type of method tends to preserve and exploit the local information in partial input.
One direct approach is to predict the missing points by combining the results with partial input data~\citep{huang2020pf,yu2021pointr}. As the point set can be viewed as a token sequence, PoinTr~\citep{yu2021pointr} and its following works~\citep{10232862,li2023proxyformer,chen2023anchorformer} employ the transformer architecture~\citep{vaswani2017attention} to predict the missing point proxies.
SeedFormer~\citep{zhou2022seedformer} introduces a shape representation called patch seeds for preventing the loss of local information during pooling operation. 
Some other approaches~\citep{pan2020ecg,10106495,zhang2022point,chen2019multi} propose to enhance the generated shapes by exploiting the structural relations in the refinement stage.
A recent work~\citep{DBLP:journals/ijcv/ZhangLXNZTL23} shares a similar idea with our \emph{Similarity Alignment} unit of utilizing self-similarities property and directly learns geometric transformations for input points.
However, these strategies employ a unified refinement strategy for all points, which limits their ability to generate pleasing details for different points.
Our approach differs from theirs by breaking down the shape refinement task into two sub-goals, and adaptively extracting reliable features for different partial areas.

\subsection{View-based Methods for 3D Understanding}
View-based 3D understanding techniques have gained significant attention in recent years.
The classic Multi-View Convolutional Neural Network (MVCNN) model was introduced in \citep{su2015multi}, where color images are fed into a CNN and subsequently combined by a pooling operation. However, this approach has the fundamental drawback of ignoring view relations.
Following works~\citep{feng2018gvcnn,wei2022learning,han20193d2seqviews,yang2019learning,chen2022imlovenet} propose various strategies to tackle this problem. For example,
\cite{yang2019learning} obtains a discriminative 3D object representation by modeling region-to-region relations.
LSTM is also used to build the inter-view relations~\citep{dai2018siamese}.
Since the cross-modal data are more available, methods are proposed to fuse features of views and point clouds.

However, compared to their 2D counterpart, most of 3D recognition methods are limited by the scarcity of training data. Therefore, in recent years, there has been a surge in research focusing on enhancing 3D understanding through the utilization of pre-trained 2D models.
P2P~\citep{p2p} prompts a pre-trained image model by the geometry-preserved projection.
PointCLIP~\citep{zhang2022pointclip} and its following work~\citep{PointCLIPV2} achieve zero-shot point cloud understanding by sending multi-view 2D projections to CLIP~\citep{radford2021learning} for a well-learned representation. To adapt the power of CLIP to more 3D scenarios, alternative methods~\citep{CLIP2,CLIP2point,clip2scene} employ cross-modal contrastive learning to train a powerful 3D encoder.
Recent approaches explore masked autoencoders using both 3D and 2D data~\citep{jointMAE, I2PMAE}. Some approaches pre-train a 3D encoder through color image generation~\citep{takeaphoto} or neural rendering~\citep{ponder}.
Inspired by the success of view-based 3D understanding techniques, our method utilizes point cloud features to enhance relationships between multiple views obtained by self-augmentation.

\subsection{Semantic Scene Completion}
Given an incomplete observation, semantic scene completion (SSC) aims to reconstruct and assign semantic labels to a 3D scene.
The prevailing approach in SSC involves utilizing the voxel grid as the primary 3D representation. Pioneering the field, SSCNet~\citep{song2017semantic} addresses this task through an end-to-end network constructed with 3D Convolutional Neural Networks. Subsequent methodologies enhance this framework by incorporating group convolution~\citep{zhang2018efficient}, multimodal fusion~\citep{guo2018view}, GAN~\citep{wang2019forknet}, and boundary learning~\citep{chen20203d}. 
SISNet~\citep{cai2021semantic} tackles SSC by iteratively performing grid-based scene completion and point cloud object completion.
AdaPoinTr~\citep{10106495} introduces a geometry-enhanced SSC framework to augment existing voxel-based SSC methods with a point cloud completion network.
To accommodate diverse autonomous driving scenes, recent research has primarily focused on outdoor LiDAR datasets and explored various input modalities.
LiDAR-based approaches~\citep{yan2021sparse,xia2023scpnet} commonly integrate point-wise features within voxel space to enhance scene understanding. Meanwhile, alternative methods have emerged to predict scene occupancy using monocular images. The pioneering work MonoScene~\citep{cao2022monoscene} addresses this problem with successive U-Nets and a novel feature projection module. Building on this foundation, \cite{li2023voxformer} introduces a two-stage transformer architecture. 
NDC-Scene~\citep{yao2023ndc} further improves performance by reducing feature ambiguity in the normalized device coordinates space. Symphonies~\citep{jiang2024symphonize} advances this line of research by incorporating instance-centric semantics and scene context through the use of instance queries.
More recently, images captured by multiple cameras are also utilized for accurate 3D understanding~\citep{Wei_2023_ICCV,huang2023tri}.
Different from these methodologies, point-based SSC methods~\citep{zhang2021point,wang2022learning,xu2023casfusionnet,yan2023pointssc} focus on reconstructing a complete scene point cloud along with semantic labels assigned to each point.
\cite{wang2022learning} attach an extra segmentation network to the point cloud completion model.
CasFusionNet~\citep{xu2023casfusionnet} takes a hierarchical approach, merging features from completion and segmentation modules.
In this work, we extend PointSea to SSC by jointly predicting semantic labels and geometric details through the proposed SDG.

\begin{figure*}[h]
  \centering
  \includegraphics[width=0.95\textwidth]{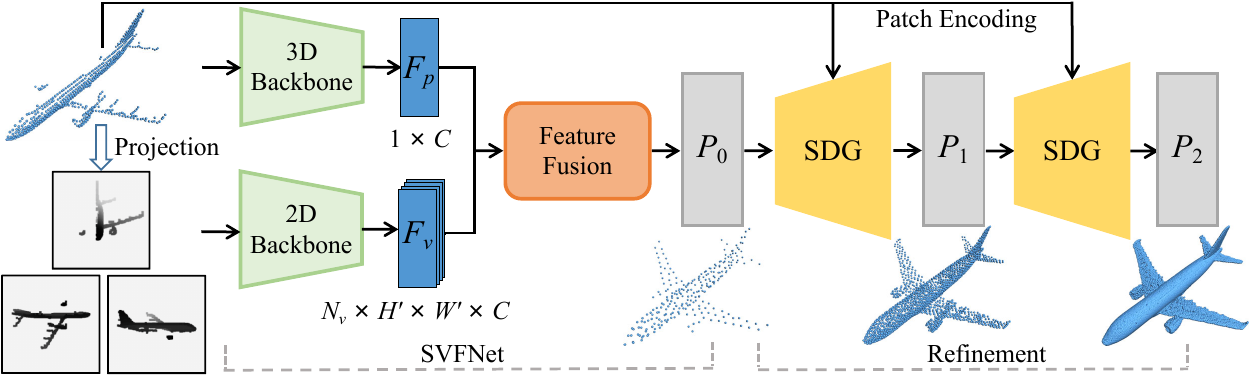}
\caption{The architecture of PointSea. SVFNet first generates a global shape from the cross-modal input. The coarse completion is then upsampled and refined with two SDGs.}
  \label{fig:overview}
\end{figure*}

\label{secRelatedWork}
\section{Method}
\begin{figure*}[h]
  \centering
  \includegraphics[width=0.96\textwidth]{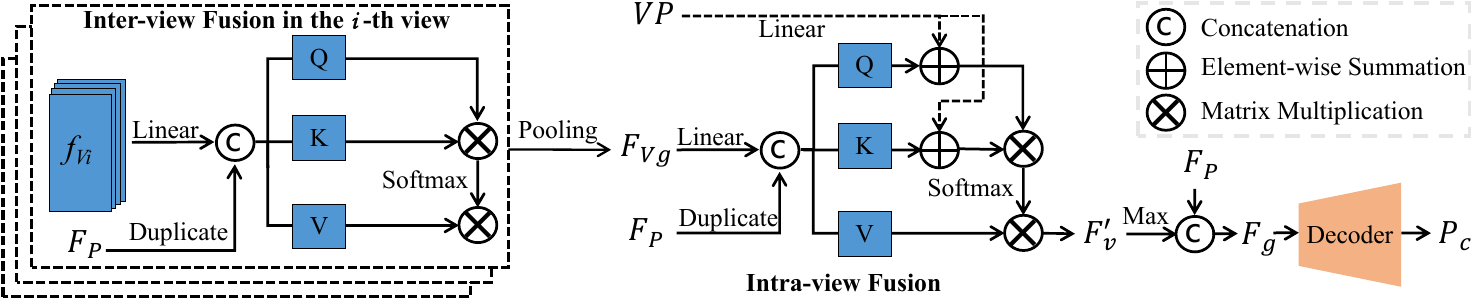}
\caption{Illustration of feature fusion module. The features from cross-modal input are fused at both inter-view and intra-view levels.}
  \label{fig:VA}
\end{figure*}
The input of our PointSea consists of three parts: a partial and low-res point cloud $P_{in}\subseteq\mathbb{R}^{N\times3}$, $N_V$ camera locations $VP\subseteq\mathbb{R}^{N_V\times3}$ (three orthogonal views in our experiments), and $N_V$ depth maps $D\subseteq\mathbb{R}^{N_V\times1\times H \times W}$.
Given these inputs, our goal is to estimate a complete point cloud $P_{2}\subseteq\mathbb{R}^{N_2\times3}$ in a coarse-to-fine manner.
The overall architecture is exhibited in Fig.~\ref{fig:overview}, which comprises two parts: an SVFNet and a refiner equipped with two SDG modules. 
SVFNet first leverages multiple self-projected depth maps to produce a globally completed shape $P_0\subseteq\mathbb{R}^{N_0\times3}$. Subsequently, two SDGs gradually refine and upsample $P_0$ to yield the final point cloud $P_{2}$, which exhibits geometric structures with high levels of detail.
Note that unlike some recent cross-modal approaches~\citep{zhang2021view,zhu2023csdn,aiello2022crossmodal,zhang2022shape}, our method makes full use of self-structures and does not require any additional paired information such as color images with precisely calibrated camera intrinsic parameters~\citep{zhang2021view,zhu2023csdn}. 
Depth maps are obtained by projecting point clouds themselves from controllable viewpoints. In contrast to the vanilla perspective projection used in SVDFormer~\citep{Zhu_2023_ICCV}, we adopt the dense projection method proposed in \citep{PointCLIPV2}, which generates smooth depth values.

\subsection{SVFNet}
\subsubsection{Overview}
The SVFNet aims to observe the partial input from different viewpoints and learns an effective descriptor to produce a globally plausible and complete shape. We first extract a global feature $F_\emph{P}$ from $P_{in}$ using a point-based 3D backbone network and a set of view features $F_V\subseteq\mathbb{R}^{N_V\times H^{\prime} \times W^{\prime} \times C}$ from the $N_V$ depth maps using a CNN-based 2D backbone network, where $H^{\prime}=H/32$ and $W^{\prime}=W/32$. 

However, how to effectively fuse the above cross-modal features is challenging.
In our early experiments, we directly concatenate these features, but the produced shape is less pleasing (see the ablation studies in Section~\ref{ablationSec}). This may be caused by the domain gap between 2D and 3D representations.
To resolve this problem, we propose a new feature fusion module, to fuse $F_\emph{P}$ and $F_V$, and output a global shape descriptor $F_g$, followed by a decoder to generate the global shape $P_c$. The decoder uses a 1D Conv-Transpose layer to transform $F_g$ to a set of point-wise features and regresses 3D coordinates with a self-attention layer~\citep{vaswani2017attention}.
Finally, we follow a similar approach to previous studies~\citep{9928787,zhou2022seedformer} by merging $P_c$ and $P_{in}$ and then resampling the merged output to generate the coarse result $P_0$.

\subsubsection{Feature Fusion}
Fig.~\ref{fig:VA} illustrates the detailed process of feature fusion. In SVDFormer~\citep{Zhu_2023_ICCV}, only intra-view fusion was conducted. In our method, the fusion process is divided into two stages.
First, we enhance the view-wise features by injecting 3D shape information, where the patch-wise features of the $i$-th view, denoted as  $f_{Vi}\subseteq\mathbb{R}^{H^{\prime} \times W^{\prime} \times C}$, are independently fused with $F_\emph{P}$.
Drawing inspiration from the vision transformer~\citep{dosovitskiy2021an}, we treat the patched features $f_{Vi}$ as a set of tokens with a length of $H^{\prime} \times W^{\prime}$. Specifically, $f_{Vi}$ is transformed into query, key, and value tokens via linear projection under the guidance of the global shape feature $F_\emph{P}$. We then calculate attention weights and perform an elemental-wise product to obtain refined features and apply maximum pooling along the spatial dimensions to derive the view-wise feature $F_{Vg}\subseteq\mathbb{R}^{N_V \times C}$. 
By fusing 3D shape information into the local 2D patches, $F_{Vg}$ serves as a good initialization for the subsequent intra-view fusion.
The second fusion is similar to the first stage, with the main difference being that, in the intra-view stage, the fusion takes place among $N_V$ views. Besides, to enhance the discriminability of view features, attention weights are calculated based on the query and key tokens conditioned on the projected viewpoints $VP$. We map $VP$ into the latent space through a linear transformation and use them as positional signals. Following an elemental-wise product, each feature in $F^\prime_V$ incorporates the relational information from other views under the guidance of $F_\emph{P}$.
Finally, the output shape descriptor $F_g$ is derived from $F^\prime_V$ via max pooling.

\subsection{SDG}
\begin{figure*}[h]
  \centering
  \includegraphics[width=\textwidth]{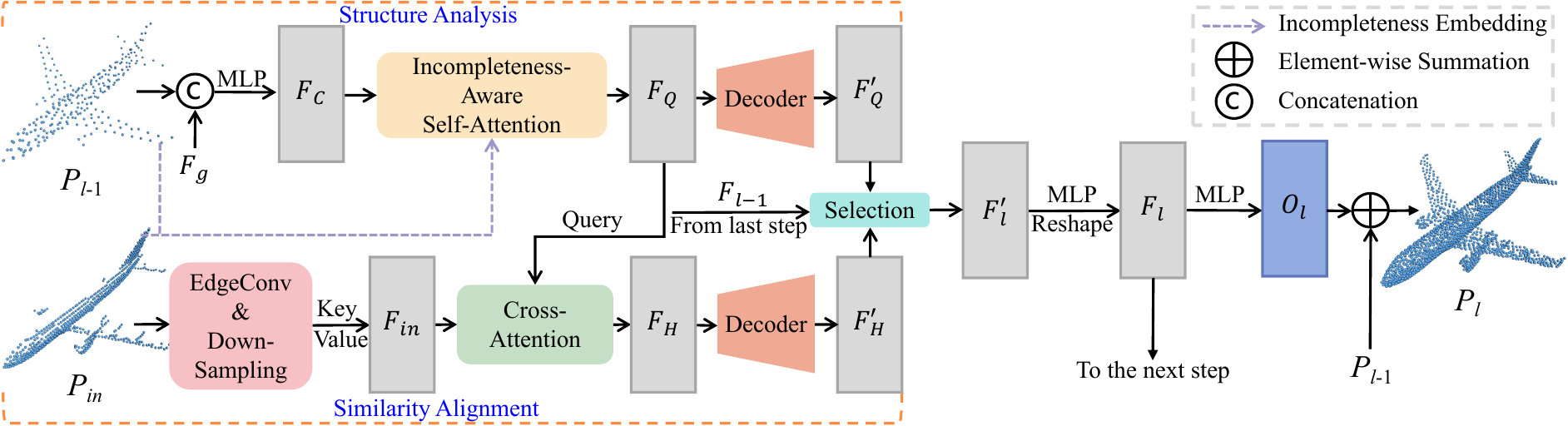}
\caption{The architecture of SDG. The upper path represents Structure Analysis and the lower path represents Similarity Alignment. Each sub-network generates an offset feature which is then combined using a Path Selection module and used to regress into the coordinate offsets.}
  \label{fig:IDTr}
\end{figure*}

The SDG seeks to generate a set of coordinate offsets to fine-tune and upsample the coarse shape, based on the structural type of the missing surface region.
To achieve it, SDG is designed as a dual-path architecture as shown in Fig.~\ref{fig:IDTr}, which consists of two parallel units named \emph{Structure Analysis} and \emph{Similarity Alignment}, respectively. 
Overall, fed with the partial input $P_{in}$, coarse point cloud $P_{l-1}$, and the offset feature $F_{l-1}$ from the previous step, we obtain a point-wise feature $F_{l}^{\prime}$. 
Denoting $r$ as the upsampling rate, $F_{l}^{\prime}$ is then projected to a higher dimensional space and reshaped to the offset feature $F_{l}\subseteq\mathbb{R}^{rN\times C}$, which will be sent to the next SDG.
Finally, we produce a set of up-sampled offsets $O_{l}\subseteq\mathbb{R}^{rN\times 3}$ through an MLP and further add it back to $P_{l-1}$ to obtain a new completion result. It is noteworthy that the SDG iteration is conducted twice, as illustrated in Fig.~\ref{fig:overview}.

\begin{figure}[h]
  \centering
  \includegraphics[width=\linewidth]{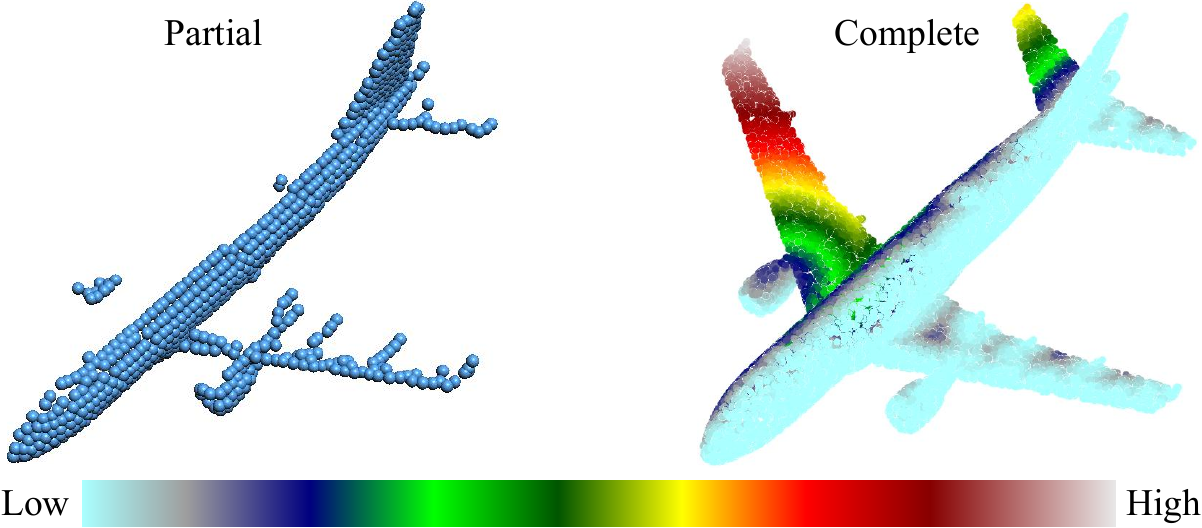}
\caption{Visualization of incompleteness degree of a reconstructed point cloud and its corresponding partial observation.}
  \label{fig:Vis_incom}
\end{figure}

\subsubsection{Structure Analysis}
Since detailed geometries from missing regions are harder to recover, we embed an incompleteness-aware self-attention layer to explicitly encourage the network to focus more on the missing regions. Specifically, $P_{l-1}$ is first concatenated with the shape descriptor $F_g$, and then embedded into a set of point-wise feature $F_C=\{f_i\}^{N_{l-1}}_{i=1}$ by a linear layer. 
Next, $F_C$ is fed to the incompleteness-aware self-attention layer to obtain a set of features $F_Q=\{q_i\}^{N_{l-1}}_{i=1}$, which encodes the point-wise incompleteness information. 
$q_i$ is computed by:
\begin{equation} \label{eqn1}
  \begin{split}
   q_i &= \sum_{j=1}^{N_{l-1}}{a_{i,j}(f_jW_V)},\\
   a_{i,j} &= Softmax((f_iW_Q+h_i)(f_jW_K+h_j)^T),
  \end{split}
\end{equation}
where $W_Q$, $W_K$, and $W_V$ are learnable matrix with the size of $C \times C$. $h_i$ is a vector that represents the degree of incompleteness for each point $x_i$ in $P_{l-1}$. 
Intuitively, points in missing regions tend to have a larger distance value to the partial input. 
Thus, we define the incompleteness degree of a point as its distance to the closest point in the partial input. We visualize the incompleteness of a reconstructed point cloud in Fig.~\ref{fig:Vis_incom}, which indicates that points in the missing areas typically have large incompleteness values. This information is embedded as follows:
\begin{equation} \label{eqn2}
  \begin{split}
   h_i &= Sinusoidal(\frac{1}{\gamma}\min_{y \in P_{in}}\lvert \lvert x_i - y \lvert\lvert)
  \end{split},
\end{equation}
where $\gamma$ is a scaling coefficient. We set it to 0.2 in our experiment and use the sinusoidal function~\citep{vaswani2017attention} to ensure that $h_i$ has the same dimension as the embeddings of query, key, and value. We then decode $F_Q$ into $F_{Q}^{\prime}$ for further analysis of the coarse shape.

\subsubsection{Similarity Alignment}
The \emph{Similarity Alignment} unit exploits the potential similar local pattern in $P_{in}$ for each point in $P_{l-1}$. 
Inspired by the point proxies in \cite{yu2021pointr}, we begin by using three EdgeConv layers~\citep{wang2019dynamic} to extract a set of downsampled point-wise feature $F_{in}$. Each vector in $F_{in}$ captures the local context information.
Given the potential existence of long-range similar structures, we then perform feature exchange through cross-attention, a classical solution for feature alignment.
The calculation process is similar to vanilla self-attention with the query matrix produced by $F_Q$ and $F_{in}$ serving as the key and value vectors.
The cross-attention layer produces point-wise features denoted as $F_H\subseteq\mathbb{R}^{N_{l-1}\times C}$, which integrates similar local structures in $P_{in}$ for each point in the coarse shape $P_{l-1}$. 
In this way, this unit can model the geometric similarity between two point clouds, thereby facilitating the refinement of points with similar structures in the input. 
Similar to the structure analysis unit, $F_H$ is also decoded into a new feature $F_{H}^{\prime}$. Both decoders have the same architecture implemented with self-attention layers~\citep{vaswani2017attention}.

\subsubsection{Path Selection}
There are two rough kinds of sources of shape information in the dual-path design: one derived from learned shape priors and the other from the similar geometric patterns found within $P_{in}$. 
In SVDFormer~\citep{Zhu_2023_ICCV}, the point-wise feature $F_{l}^{\prime}$ was obtained through the concatenation of features from the two paths. 
However, the significance of these two units varies across different missing regions. For instance, in certain missing regions, the input observation exhibits nearly identical geometry due to properties such as symmetry. In such scenarios, information derived from the \emph{Similarity Alignment} unit plays a more significant role for these points. Motivated by this observation, this paper integrates these two feature types via a path selection module, which adaptively chooses the more crucial feature. 

This module is designed to dynamically select more essential information for each point. The core idea involves controlling information flow through gates that integrate global context from both the current and previous refinement steps.
Specifically, $F_{l}^{\prime}$ is computed by:
\begin{equation} \label{eqn3}
  \begin{split}
   F_{l}^{\prime} &= \alpha F_{Q}^{\prime} + (1-\alpha) F_{H}^{\prime},\\
   \alpha &= Sigmoid(MLP([F_{Q}^{\prime}+F_{H}^{\prime},F_{l-1},Max(F_Q)])),
  \end{split}
\end{equation}
where $F_{l-1}$ is the offset feature from the last SDG.
The efficacy of connecting multiple refinement steps for integrating spatial relationships has been consistently demonstrated in prior works~\citep{9928787,zhou2022seedformer,yan2022fbnet}. In our approach, we intend to leverage the spatial splitting information from the previous step to guide the selection of different features in the current step.
We also apply max pooling to $F_Q$ and utilize the incompleteness-aware global feature to assist the selection.
Note that in the first SDG, $F_{l-1}$ is not involved in the path selection module due to its absence.

\subsection{Loss Function}
To measure the differences between the generated point cloud and the ground truth $P_{gt}$, we use the Chamfer Distance (CD) as our loss function, which is a common choice in recent works. 
To facilitate the coarse-to-fine generation process, we regularize the training by computing the loss function as:
\begin{equation} \label{eqnloss}
    \mathcal{L}= \mathcal{L}_{CD}(P_c,P_{gt}) + \sum_{i = {1,2}}\mathcal{L}_{CD}(P_i,P_{gt}),
\end{equation}
where $\mathcal{L}_{CD}$ is defined as
\begin{equation}\label{eqnCD}
    \mathcal{L}_{CD}(X,Y)=\frac{1}{\lvert X \rvert}\sum_{x \in X}\min_{y \in Y} \lvert \lvert x-y \rvert \rvert + \frac{1}{\lvert Y \rvert}\sum_{y \in Y}\min_{x \in X} \lvert \lvert y-x \rvert \rvert.
\end{equation}
Note that we downsample the $P_{gt}$ to the same density as $P_c$, $P_1$, and $P_2$ to compute the corresponding losses.

\label{secOverview}

\section{Experiment}
In this section, we first conduct experiments on the widely-used completion benchmarks: ShapeNet-55/34~\citep{10232862}, PCN~\citep{yuan2018pcn}, and Projected-ShapeNet-55/34~\citep{10232862}. 
These datasets leverage various techniques to generate incomplete point clouds, allowing for a comprehensive assessment of PointSea.
Following this, we evaluate our method on real-world partial scans sourced from KITTI~\citep{geiger2013vision}, ScanNet~\citep{dai2017scannet}, and Matterport3D~\citep{matterport3d}.
Additionally, we extend our method to point cloud semantic scene completion, conducting experiments on two indoor datasets~\citep{xu2023casfusionnet}.
Finally, a series of ablation studies are performed to demonstrate the impact of each component.
\begin{figure*}[t]
  \centering
  \includegraphics[width=\textwidth]{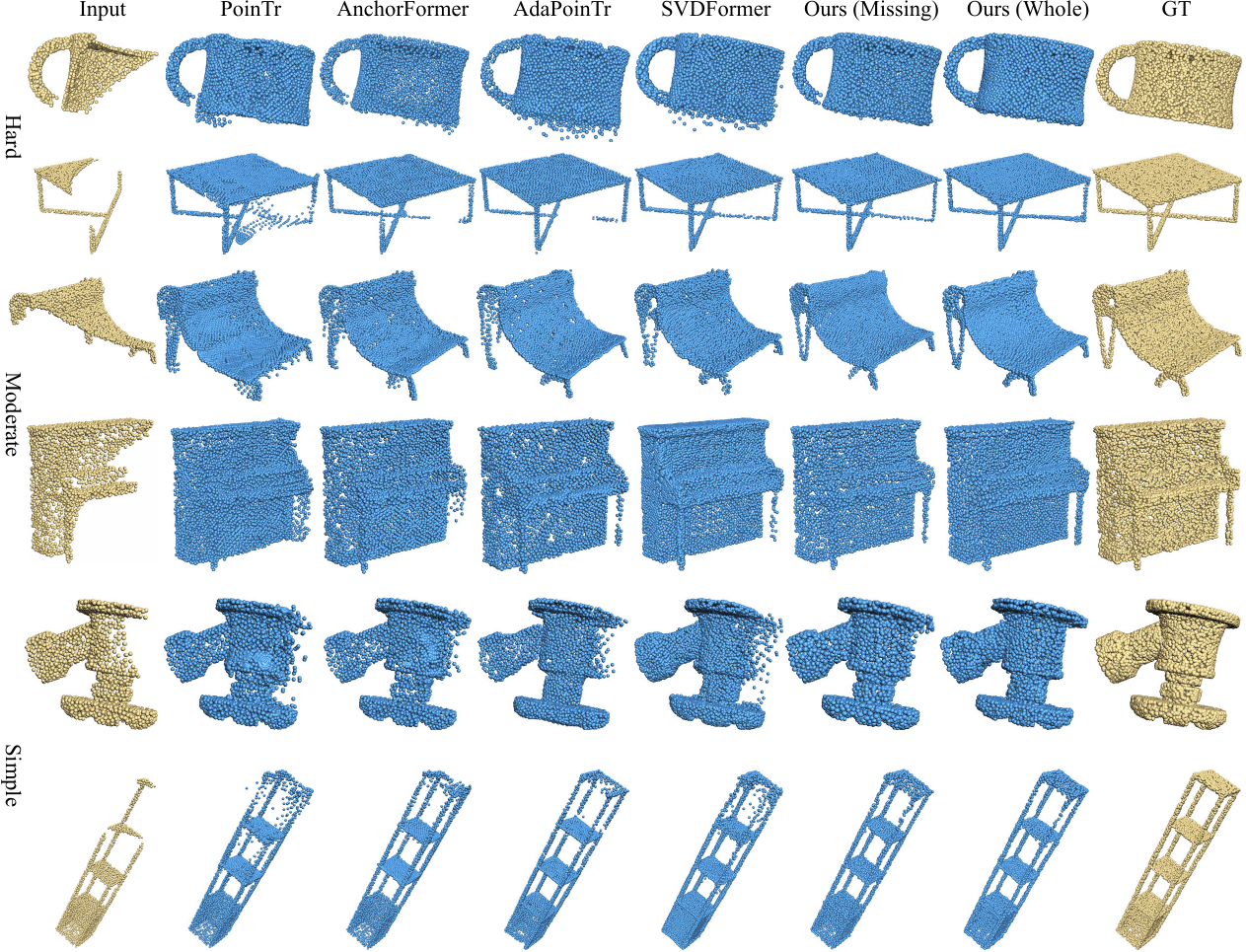}
\caption{Visual comparison with recent methods~\citep{yu2021pointr,chen2023anchorformer,10232862,Zhu_2023_ICCV} on ShapeNet-55. Hard,  Moderate, and Simple stand for the three difficulty levels.}
  \label{fig:vis55}
\end{figure*}

\begin{table*} 
    \renewcommand\arraystretch{1.2}
    \centering
    \caption{Quantitative results on ShapeNet-55. CD-S, CD-M, and CD-H stand for CD values under the simple, moderate, and hard difficulty levels, respectively ({$\displaystyle \ell ^{2}$} CD $\times 10^3$, DCD, and F1-Score@1\%). Ours (Whole) denotes PointSea, which generates 8,192 points as the completion results. On the other hand, Ours (Missing) denotes PointSea that produces 6,144 points and combines them with the partial input as the final prediction.}
    \small
    \label{tab:shapenet55}
    \begin{tabular}{|c|ccc|ccc|}
    \hline
    Methods & CD-S & CD-M & CD-H & CD-Avg$\downarrow$ & DCD-Avg$\downarrow$ & F1$\uparrow$ \\
    \hline
              \multicolumn{7}{|c|}{Methods that predict the whole shape}\\
              \hline
              FoldingNet~\citep{yang2018foldingnet} & 2.67 & 2.66 & 4.05 & 3.12 &0.826&  0.082\\
              PCN~\citep{yuan2018pcn} & 1.94 & 1.96 & 4.08 & 2.66 & 0.618  & 0.133\\
              TopNet~\citep{tchapmi2019topnet} & 2.26 & 2.16 & 4.3 & 2.91 &0.743&0.126 \\
              GRNet~\citep{xie2020grnet}  & 1.35 & 1.71 & 2.85 & 1.97 &0.592&0.238 \\
              SeedFormer~\citep{zhou2022seedformer} & 0.50 & 0.77 & 1.49 & 0.92 &0.558& 0.472 \\
              GTNet~\citep{DBLP:journals/ijcv/ZhangLXNZTL23}  & 0.45 & 0.66 & 1.30 & 0.80 & 0.696 & 0.543 \\
              AdaPoinTr~\citep{10232862}  & 0.49 & 0.69 & 1.24 & 0.81 & 0.635 & 0.503 \\
              SVDFormer~\citep{Zhu_2023_ICCV} &  0.48 & 0.70 & 1.30 & 0.83 & 0.541 & 0.451 \\

              \hline
              \multicolumn{7}{|c|}{Methods that predict the missing part}\\
              \hline
              PFNet~\citep{huang2020pf} & 3.83 & 3.87 & 7.97 & 5.22 & 0.716 &0.339 \\
              PoinTr~\citep{yu2021pointr}  & 0.58 & 0.88 & 1.79 & 1.09 &0.575& 0.464 \\
              ProxyFormer~\citep{li2023proxyformer}  & 0.49 & 0.75 & 1.55 & 0.93 & 0.549 & 0.483 \\
              AnchorFormer~\citep{chen2023anchorformer}  & 0.41 & 0.61 & 1.26 & 0.76 & 0.584 & 0.558 \\
              
              \hline
              Ours (Whole) &  0.43 & 0.64 & 1.19 & 0.75
               & \textbf{0.532} & 0.485 \\
              Ours (Missing) &  \textbf{0.35}  & \textbf{0.57}  & \textbf{1.18}  & \textbf{0.70}  & 0.563  & \textbf{0.564} \\
              \hline
    \end{tabular}
\end{table*}

\begin{table*}[htb]
    \renewcommand\arraystretch{1.2}
    \centering
    \caption{Quantitative results on ShapeNet-34. CD-S, CD-M, and CD-H stand for CD values under the simple, moderate, and hard difficulty levels, respectively ({$\displaystyle \ell ^{2}$} CD $\times 10^3$, DCD, and F1-Score@1\%).}
    \footnotesize
    \label{tab:shapenet34}
    \begin{tabular}{|c|cccccc|cccccc|}
    \hline
    \multirow{2}{*}{Methods} &  \multicolumn{6}{c}{34 seen categories} & \multicolumn{6}{c}{21 unseen categories} \\ \cline{2-13} & \makebox[0.01\linewidth][c]{CD-S} & \makebox[0.01\linewidth][c]{CD-M} & \makebox[0.01\linewidth][c]{CD-H} & \makebox[0.05\linewidth][c]{CD-Avg$\downarrow$} & \makebox[0.05\linewidth][c]{DCD-Avg$\downarrow$} & \makebox[0.01\linewidth][c]{F1$\uparrow$} & \makebox[0.01\linewidth][c]{CD-S} & \makebox[0.01\linewidth][c]{CD-M} & \makebox[0.01\linewidth][c]{CD-H} & \makebox[0.05\linewidth][c]{CD-Avg$\downarrow$} & \makebox[0.05\linewidth][c]{DCD-Avg$\downarrow$} & \makebox[0.01\linewidth][c]{F1$\uparrow$} \\
    \hline
    \multicolumn{13}{|c|}{Methods that predict the whole shape}\\
    \hline
    FoldingNet~\citep{yang2018foldingnet} & 1.86 & 1.81 & 3.38 & 2.35 & 0.831 & 0.139 & 2.76 & 2.74 & 5.36 & 3.62 & 0.870 & 0.095\\
    PCN~\citep{yuan2018pcn} & 1.87 & 1.81 & 2.97 & 2.22 & 0.624 & 0.150 & 3.17 & 3.08 & 5.29 & 3.85 & 0.644 & 0.101\\
    TopNet~\citep{yuan2018pcn} & 1.77 & 1.61 & 3.54 & 2.31 & 0.838 & 0.171 & 2.62 & 2.43 & 5.44 & 3.50 & 0.825 & 0.121\\
    GRNet~\citep{xie2020grnet} & 1.26 & 1.39 & 2.57 & 1.74 & 0.600 & 0.251 & 1.85 & 2.25 & 4.87 & 2.99 & 0.625 & 0.216\\
    SeedFormer~\citep{zhou2022seedformer} & 0.48 & 0.70 & 1.30 & 0.83 & 0.561 & 0.452 & 0.61 & 1.07 & 2.35 & 1.34 & 0.586 & 0.402\\
    GTNet~\citep{DBLP:journals/ijcv/ZhangLXNZTL23} & 0.51 & 0.73 & 1.40 & 0.88 & 0.700 & 0.511 & 0.78 & 1.22 & 2.56 & 1.52 & 0.703 & 0.467\\
    AdaPoinTr~\citep{10232862} & 0.48 & 0.63 & 1.07 & 0.73 & 0.606 & 0.469 & 0.61 & 0.96 & 2.11 & 1.23 & 0.619 & 0.416\\
    SVDFormer~\citep{Zhu_2023_ICCV} & 0.46  & 0.65 & 1.13  & 0.75 & 0.538 & 0.457 & 0.61 & 1.05 & 2.19 & 1.28 & 0.554 & 0.427\\
    
    \hline
    \multicolumn{13}{|c|}{Methods that predict the missing part}\\
    \hline
    
    PFNet~\citep{huang2020pf} & 3.16 & 3.19 & 7.71 & 4.68 & 0.708 & 0.347 & 5.29 & 5.87 & 13.33 & 8.16 & 0.723 & 0.322\\
    PoinTr~\citep{yu2021pointr} & 0.76 & 1.05 & 1.88 & 1.23 & 0.575 & 0.421 & 1.04 & 1.67 & 3.44 & 2.05 & 0.604 & 0.384\\
    ProxyFormer~\citep{li2023proxyformer} & 0.44 & 0.67 & 1.33 & 0.81 & 0.556 & 0.466 & 0.60 & 1.13 & 2.54 & 1.42 & 0.583 & 0.415\\
    AnchorFormer~\citep{chen2023anchorformer} & 0.41 & 0.57 & 1.12 & 0.70 & 0.585 & 0.564 & 0.52 & 0.90 & 2.16 & 1.19 & 0.598 & 0.535\\
    
    \hline
    Ours (Whole) & 0.40  & 0.57 & 1.00  & 0.66 & \textbf{0.525} & 0.492 & 0.50 & 0.88 & \textbf{1.92} & 1.10 & \textbf{0.541} & 0.461\\
    Ours (Missing) & \textbf{0.32}  & \textbf{0.50} & \textbf{0.98}  & \textbf{0.60} & 0.555 & \textbf{0.575} &\textbf{0.43} & \textbf{0.83} & 1.95 & \textbf{1.07} & 0.572 & \textbf{0.548}\\
    \hline
    \end{tabular}
\end{table*}

\begin{figure*}[t]
  \centering
  \includegraphics[width=\textwidth]{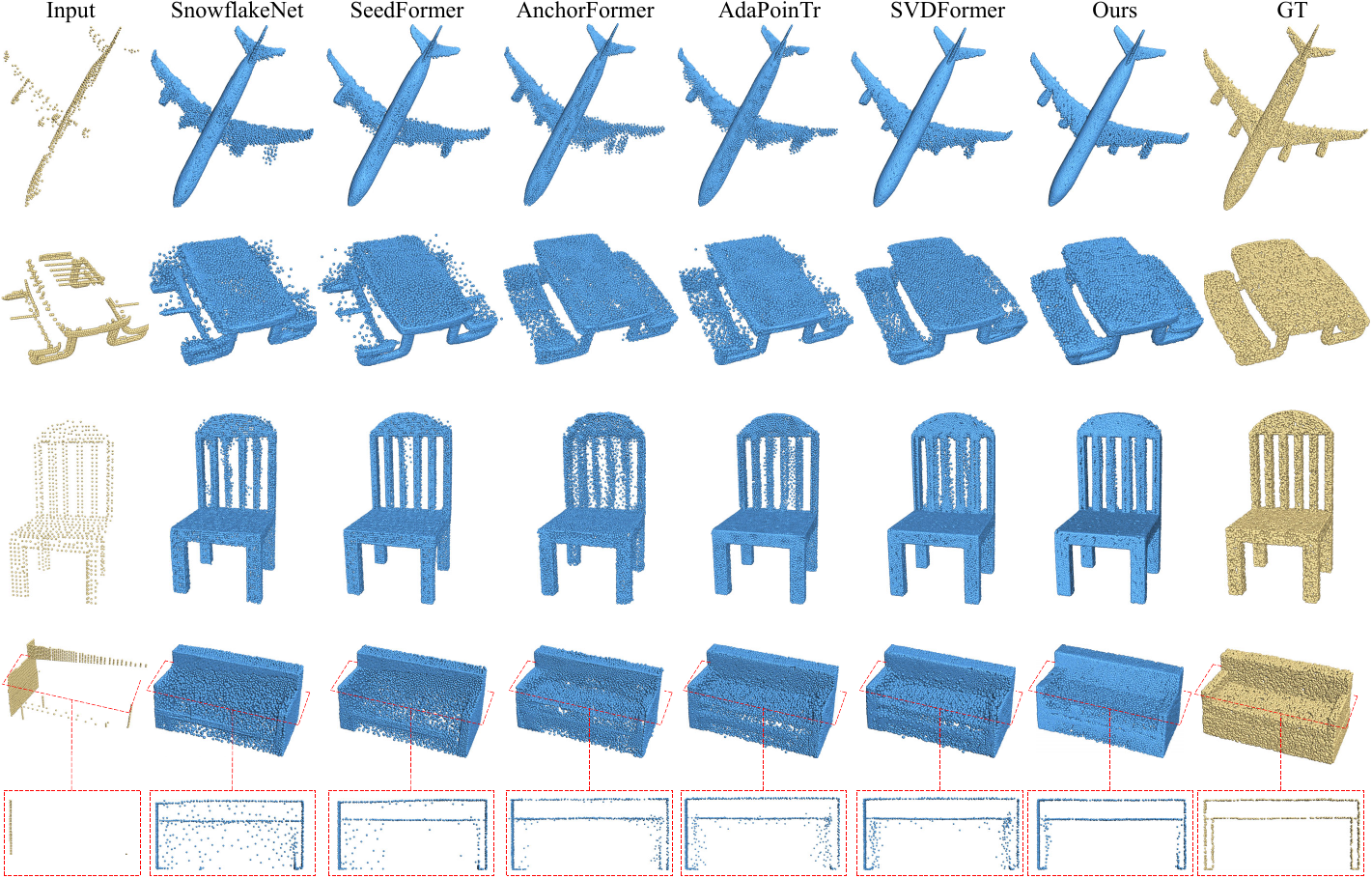}
\caption{Visual comparisons with recent methods~\citep{9928787,zhou2022seedformer,chen2023anchorformer,10232862,Zhu_2023_ICCV} on the PCN dataset. Our method produces the most faithful and detailed structures compared to its competitors. In the fourth row, we provide a visualization of a thin cross-section for the couch model, showcasing PointSea's excellence in reconstruction quality with fewer noisy points. }
  \label{fig:Vis_PCN}
\end{figure*}

\begin{table*}[t]
    \tiny
    \renewcommand\arraystretch{1.2}
    \centering
    \caption{Quantitative results on the PCN dataset. ({$\displaystyle \ell ^{1}$} CD $\times 10^3$, DCD, and F1-Score@1\%)}
    \label{tab:pcnl1}
    \small
    \begin{tabular}{|c|cccccccc|ccc|}
    \hline
    Methods & Plane & Cabinet & Car & Chair & Lamp & Couch & Table & Boat & CD-Avg$\downarrow$  & DCD$\downarrow$  &F1$\uparrow$  \\
    \hline
              PCN~\citep{yuan2018pcn} & 5.50 & 22.70 & 10.63 & 8.70 & 11.00 & 11.34 & 11.68 & 8.59 & 9.64& 0.622 &0.695\\
              GRNet~\citep{xie2020grnet}  & 6.45 & 10.37 & 9.45 & 9.41 & 7.96 & 10.51 & 8.44 & 8.04& 8.83& 0.622 &0.708\\
              CRN~\citep{wang2020cascaded}    & 4.79 & 9.97 & 8.31 & 9.49 & 8.94 & 10.69 & 7.81 & 8.05& 8.51& 0.594 & 0.652 \\
              NSFA~\citep{zhang2020detail}    & 4.76 & 10.18 & 8.63 & 8.53 & 7.03 & 10.53 & 7.35 & 7.48 & 8.06 & 0.578 & 0.734 \\
              PoinTr~\citep{yu2021pointr}  & 4.75 & 10.47 & 8.68 & 9.39 & 7.75 & 10.93 & 7.78 & 7.29 & 8.38 & 0.611 & 0.745 \\
              SnowflakeNet~\citep{9928787} & 4.29 & 9.16 & 8.08 & 7.89 & 6.07 & 9.23 & 6.55 & 6.40& 7.21& 0.585 & 0.801 \\
              SDT~\citep{zhang2022point} & 4.60 & 10.05 & 8.16 & 9.15 & 8.12 & 10.65 & 7.64 & 7.66 & 8.24 & 0.637 &0.754 \\
              PMP-Net++~\citep{wen2022pmp} & 4.39 & 9.96 & 8.53 & 8.09 & 6.06 & 9.82 & 7.17 & 6.52& 7.56 & 0.611 & 0.781\\
              SeedFormer~\citep{zhou2022seedformer}  & 3.85 & 9.05 & 8.06 & 7.06 & 5.21 & 8.85 & 6.05 & 5.85 & 6.74 & 0.583 & 0.818 \\
              FBNet~\citep{yan2022fbnet}  & 3.99 & 9.05 & 7.90 & 7.38 & 5.82 & 8.85 & 6.35 & 6.18& 6.94& - & -\\
              LAKeNet~\citep{tang2022lake}  & 4.17 & 9.78 & 8.56 & 7.45 & 5.88 & 9.39 & 6.43 & 5.98 & 7.23 & - & -\\

              VAPCNet~\citep{fu2023vapcnet}  & 4.10 & 9.28 & 8.15 & 7.51 & 5.55 & 9.18 & 6.28 & 6.10 & 7.02 & - & - \\
              GTNet~\citep{DBLP:journals/ijcv/ZhangLXNZTL23}  & 4.17 & 9.33 & 8.38 & 7.66 & 5.49 & 9.44 & 6.69 & 6.07 & 7.15 & - & - \\
              CP3~\citep{10097548}  & 4.34 & 9.02 & 7.90 & 7.41 & 6.35 & 8.52 & 6.32 & 6.26 & 7.02 & - & - \\
              AnchorFormer~\citep{chen2023anchorformer}  & 3.70 & 8.94 & 7.57 & 7.05 & 5.21 & 8.40 & 6.03 & 5.81 & 6.59 & 0.554 & 0.827 \\
              AdaPoinTr~\citep{10232862}  & 3.68 & 8.82 & 7.47 & 6.85 & 5.47 & 8.35 & 5.80 & 5.76 & 6.53 & 0.538 & 0.845 \\
              SVDFormer~\citep{Zhu_2023_ICCV} & 3.62 & 8.79 & 7.46 & 6.91 & 5.33 & 8.49 & 5.90 & 5.83& 6.54 & 0.536 & 0.841 \\

              \hline
              Ours & \textbf{3.52} & \textbf{8.54} & \textbf{7.33} & \textbf{6.58} & \textbf{5.21} & \textbf{8.24} & \textbf{5.75} & \textbf{5.62}& \textbf{6.35} & \textbf{0.522} &\textbf{0.858} \\
              \hline
    \end{tabular}
\end{table*}

\begin{table*}[htb]
    \renewcommand\arraystretch{1.1}
    \centering
    \caption{Quantitative results on Projected-ShapeNet-55/34. ({$\displaystyle \ell ^{1}$} CD $\times 10^3$ and F1-Score@1\%)}
    \small
    \label{tab:projectshapenet}
    \begin{tabular}{|c|cc|cc|cc|}
    \hline
    \multirow{2}{*}{Methods} &  \multicolumn{2}{c|}{55 categories} & \multicolumn{2}{c|}{34 seen categories} & \multicolumn{2}{c|}{21 unseen categories} \\ \cline{2-7} & CD$\downarrow$ & F1$\uparrow$ & CD$\downarrow$ & F1$\uparrow$ & CD$\downarrow$ & F1$\uparrow$  \\
    \hline
    PCN~\citep{yuan2018pcn} & 16.64 & 0.403 & 15.53 & 0.432 & 21.44 & 0.307\\
    TopNet~\citep{yuan2018pcn} & 16.35 & 0.337 & 12.96 & 0.464 & 15.98 & 0.358\\
    GRNet~\citep{xie2020grnet} & 12.81 & 0.491 & 12.41 & 0.506 & 15.03 & 0.439\\
    SnowflakeNet~\citep{9928787} & 11.34 & 0.594 & 10.69 & 0.616 & 12.82 & 0.551\\
    PoinTr~\citep{yu2021pointr} & 10.68 & 0.615 & 10.21 & 0.634 & 12.43 & 0.555\\
    SeedFormer~\citep{zhou2022seedformer} & 11.17 & 0.602 & 10.88 & 0.608 & 12.85 & 0.553\\
    AnchorFormer~\citep{chen2023anchorformer} & 10.34 & 0.665 & 9.79 & 0.683 & 12.10 & 0.612\\
    AdaPoinTr~\citep{10232862} & 9.58 & 0.701 & 9.12 & 0.721 & 11.37 & 0.642\\
    SVDFormer~\citep{Zhu_2023_ICCV} & 9.71 & 0.689 & 9.26 & 0.704 & 11.51 & 0.620\\
    
    \hline
    Ours &  \textbf{9.25} & \textbf{0.727} & \textbf{8.79} & \textbf{0.745} & \textbf{11.05} & \textbf{0.658} \\
    \hline
    \end{tabular}
\end{table*}

\subsection{Implementation Details}
We use the dense projection method introduced by \cite{PointCLIPV2} to generate depth maps with a resolution of 224 $\times$ 224 from three orthogonal views. These projected depth maps, without applying any color mapping enhancement, are directly fed to the network. 
For point clouds normalized to the range [-0.5, 0.5], such as those in the PCN dataset, the depth maps are projected at a distance of 0.7 to observe the entire shape. For point clouds normalized to the range [-1.0, 1.0], like those in the ShapeNet-55/34 dataset, the distance is set to 1.5.

In SVFNet, we use PointNet++~\citep{qi2017pointnet++} to extract features from point clouds. The detailed architecture is: 
$\emph{SA}(C = [3,64,128], N = 512, K = 16)\rightarrow\emph{SA}(C = [128,256], N = 128, K = 16)\rightarrow\emph{SA}(C = [512,256])$. 
We leverage a ResNet-18~\citep{he2016deep} model, pre-trained on ImageNet~\citep{5206848}, to extract features from depth maps. A self-attention layer with 512 hidden feature dimensions, followed by an MLP, is utilized to regress the coarse points $P_C$. The merged point cloud $P_0$ comprises 512 points for PCN and 1024 points for ShapeNet-55/34.

In SDG, we adopt EdgeConv~\citep{wang2019dynamic} to extract local features from $P_{in}$. The detailed architecture is:
$\emph{EdgeConv}(C = [3,64], K = 16)\rightarrow\emph{FPS}(2048,512)\rightarrow\emph{EdgeConv}(C = [64,256], K = 8)$. 
After obtaining $F_Q$ and $F_H$, we use a decoder composed of two self-attention layers (one in the ShapeNet-55/34 experiments) to further analyze the coarse shapes.
$F_{l}^{\prime}$ is then passed to an MLP and reshaped to $F_{l}\subseteq\mathbb{R}^{rN\times 128}$. Finally, the coordinates offset is predicted by an MLP with feature dimensions of [128, 64, 3].
Note that we use two SDGs in the refinement step. 
For the two SDG modules, we use the shared-weights architecture above. The hidden feature dimensions of self-attention layers are set as 768 and 512. 
The upsampling rates {$r_1$, $r_2$} are set to \{4, 8\} for PCN and \{2, 4\} for ShapeNet-55/34, respectively.

The network is implemented using PyTorch~\citep{paszke2019pytorch} and trained with the Adam optimizer~\citep{kingma2014adam} on NVIDIA 3090 GPUs. It takes 400 epochs for convergence. The initial learning rate is set to 0.0001 and decayed by 0.7 for every 40 epochs.

\subsection{Evaluation on ShapeNet-55/34}
\subsubsection{Data and Metrics}
The ShapeNet-55~\citep{yu2021pointr} dataset is created based on ShapeNet~\citep{chang2015shapenet} and contains shapes from 55 categories with 41,952 shapes for training and 10,518 shapes for testing. 
ShapeNet-34~\citep{yu2021pointr} is characterized by 34 categories for training, with 21 additional unseen categories reserved for testing. In total, 46,765 shapes from the 34 categories are used for training. Testing involves 3,400 shapes from the seen 34 categories and 2,305 shapes from the unseen 21 categories.
In both datasets, The ground-truth point cloud has 8,192 points and the partial input has 2,048 points. 
During training, to generate incomplete point clouds, we randomly select a viewpoint and remove the $n$ furthest points, with the remaining points down-sampled to 2,048. During testing, 8 fixed viewpoints are employed, and the number of missing points is set to 2,048, 4,096, and 6,144, corresponding to three difficulty levels: simple (S), moderate (M), and hard (H).
We use $\displaystyle \ell^{2}$ version of CD, Density-aware Chamfer Distance (DCD)~\citep{wu2021balanced}, and F-Score@1\% as evaluation metrics.

\subsubsection{Quantitative Results}
The quantitative results are summarized in Tab.~\ref{tab:shapenet55} and Tab.~\ref{tab:shapenet34}, consisting of CD values for three difficulty levels and the average value of two additional metrics. We organize the results of existing methods based on their completion strategies, categorizing them into two groups: those generating all 8,192 points as the final prediction~\citep{yang2018foldingnet,yuan2018pcn,tchapmi2019topnet,xie2020grnet,9928787,zhou2022seedformer,10232862,Zhu_2023_ICCV,DBLP:journals/ijcv/ZhangLXNZTL23}, and those only producing the missing 6,144 points and then combining them with the partial input to form the final results~\citep{huang2020pf,yu2021pointr,li2023proxyformer,chen2023anchorformer}. Also based on this, for our method, we provide two versions, denoted as Ours (Whole) and Ours (Missing).

Specifically, when producing the whole shape (Ours (Whole)) following the same setting as SVDFormer~\citep{Zhu_2023_ICCV}, PointSea outperforms all other methods in terms of CD and DCD, as shown in Tab.~\ref{tab:shapenet55} and Tab.~\ref{tab:shapenet34}. 
Furthermore, in comparison to SVDFormer~\citep{Zhu_2023_ICCV}, PointSea exhibits a noteworthy reduction in average CD by 9.6\% and 14.1\% on ShapeNet-55 and Unseen categories of ShapeNet-34, respectively. This signifies that the new designs not only contribute to improved completion performance but also enhance robustness to unseen data.
It is important to note that although GTNet~\citep{DBLP:journals/ijcv/ZhangLXNZTL23} and PointSea share similar concepts of utilizing similar geometries in the input, our approach demonstrates superior generalization ability. As reported in Tab.~\ref{tab:shapenet34}, our approach achieves an average CD that is 38.2\% lower than GTNet~\citep{DBLP:journals/ijcv/ZhangLXNZTL23} on the unseen categories.
Additionally, when only producing the missing 6,144 points, we can observe that PointSea demonstrates substantial performance improvement, surpassing even the whole version.

\subsubsection{Qualitative Results}
Fig.~\ref{fig:vis55} shows the visualization of results produced by different methods under all three difficulty levels. Overall, PointSea demonstrates better performance in generating the missing parts with faithful details.
Moreover, the advantage of our SDG can be well proved by the cases of models in the second, third, and fourth rows, where the missing parts have similar structures in the input. Current methods like PoinTr~\citep{yu2021pointr}, AnchorFormer~\citep{chen2023anchorformer}, and AdaPoinTr~\citep{10232862} fail to generate the complex missing geometries. Although SVDFormer completes them by imitating the input, the results contain undesired details with noisy points. With the assistance of the path selection module, PointSea goes a step further, producing cleaner results.

\subsubsection{Reasons for the Performance Gap Between Different Completion Strategies}
We observe a large performance gap between methods using different completion strategies (only missing part or the whole model) on the ShapeNet-55/34 dataset. When predicting only the missing part, PointSea performs significantly better in CD and F1-Score, while falling behind in DCD and visual performance. 
This phenomenon is primarily attributed to the fact that \textbf{the input partial point cloud is a subset of the ground-truth point cloud due to the online cropping strategy of ShapeNet-55/34}. 

The CD metric is a bi-directional measure that summarizes the distance from prediction (Pred) to Ground Truth (GT) and vice versa. When calculating the distance from Pred to GT, the method predicting the missing portion will generate a value of 0 for the input 2,048 points. In contrast, for the distance from GT to Pred, although the input is sparser than the ground-truth point cloud, the value remains close to 0 in these regions. However, when predicting the whole shape, the network might not precisely replicate the ground truth within the input area, leading to higher CD values, especially in simple and moderate settings. 
Although F1-Score@1\% is also computed based on the bi-directional distance, the difference from CD is that F1-Score strictly classifies distance values as either ``True" ($\leq 0.01$) or ``False" ($> 0.01$). 
Therefore, for the method that predicts the whole shape, points in the input region whose distance is slightly larger than the threshold of 0.01 will be regarded as ``False", thus widening the quantitative gap. Naturally, methods predicting the missing part will achieve much higher F1-Score in all three difficulty levels.

In general, only predicting the missing part allows the network to achieve better CD and F1-Score. However, employing this strategy does not ensure the consistency of the entire shape. This may result in point clouds that are unevenly distributed and exhibit discontinuities, as visualized by the mug and table models in Fig.~\ref{fig:vis55}. Some other methods, such as PoinTr~\citep{yu2021pointr} and AnchorFormer~\citep{chen2023anchorformer}, that similarly adopt the missing part prediction strategy, exhibit a similar phenomenon. 
This limitation is further emphasized by the comparison in terms of DCD. Before calculating the point-wise bi-directional distance, both CD and DCD construct a one-to-one mapping between two point sets using the nearest neighbor search strategy. However, DCD additionally introduces a normalization term into the distance calculation~\citep{wu2021balanced}.
This choice makes DCD more sensitive to the local point density. As observed in Tab.~\ref{tab:shapenet55}, the DCD of PointSea increases from 0.532 to 0.563 when predicting the missing part.
Considering that in most cases, the input does not constitute a subset of the ground-truth point cloud and may even contain substantial noise~\citep{10232862}. So, in the following experiments, we choose the whole shape prediction strategy as the default setting of our method, enabling the network to adaptively learn to complete the input point clouds.

\subsection{Evaluation on PCN}
\label{secPCN}
\subsubsection{Data and Metrics}
The PCN dataset~\citep{yuan2018pcn} contains shapes of 8 categories in ShapeNet~\citep{chang2015shapenet}. The ground-truth point cloud has 16,384 points and the partial input has 2,048 points. The ground-truth point cloud consists of 16,384 points, while the partial input comprises 2,048 points. Ground-truth point clouds are uniformly sampled from the shape surface. Incomplete point clouds are generated by back-projecting 2.5D depth images from 8 viewpoints. We follow the same experimental setting as previous works~\citep{yuan2018pcn,9928787,10232862,zhou2022seedformer} for a fair comparison. The $\displaystyle \ell^{1}$ version of CD, DCD, and F-Score@1\% are used as evaluation metrics.

\begin{figure}[h]
  \centering
  \includegraphics[width=\linewidth]{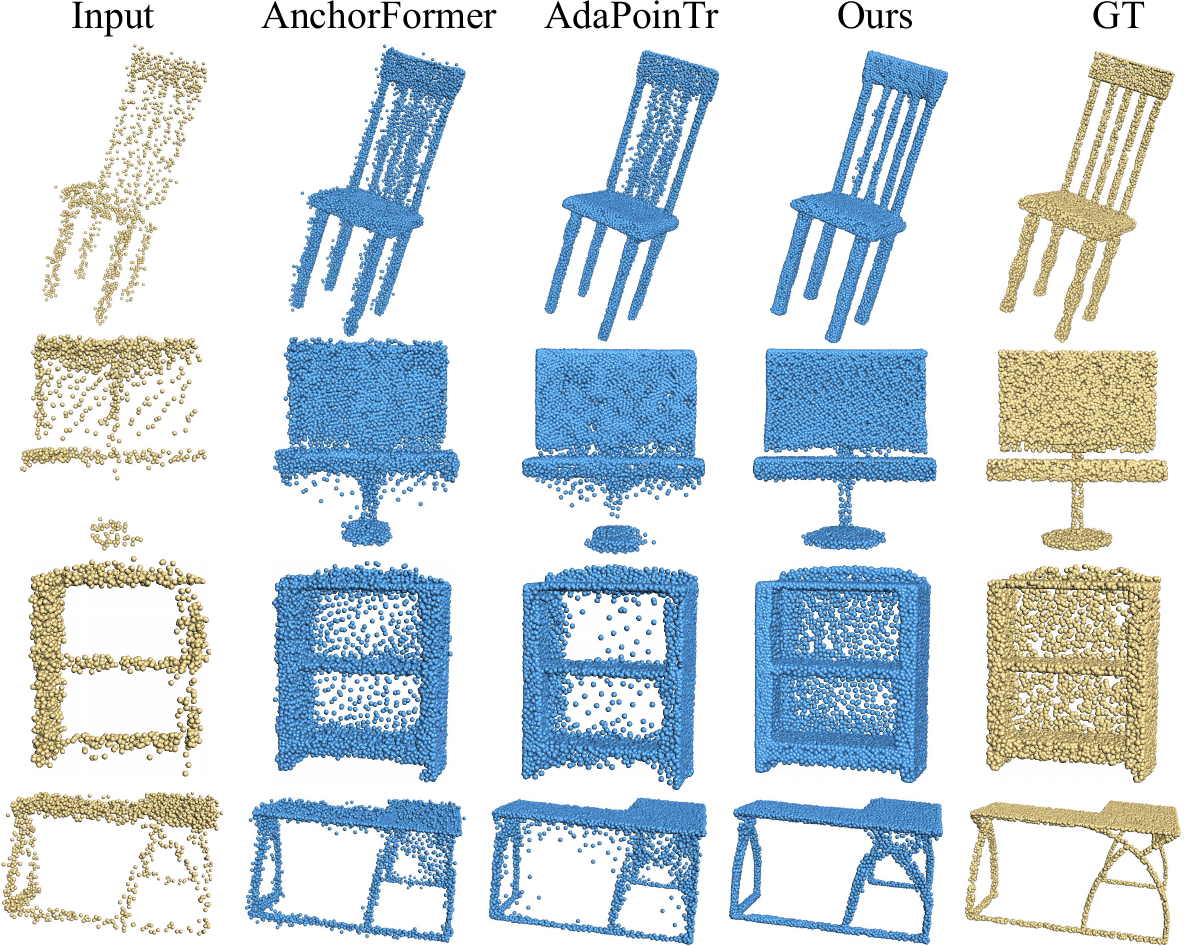}
\caption{Visual comparisons with recent methods~\citep{chen2023anchorformer,10232862} on the Projected-ShapeNet-55 dataset. When the partial input contains substantial noisy points, PointSea demonstrates the ability to refine these regions.}
  \label{fig:Vis_p55}
\end{figure}

\begin{figure*}[h]
  \centering
  \includegraphics[width=\textwidth]{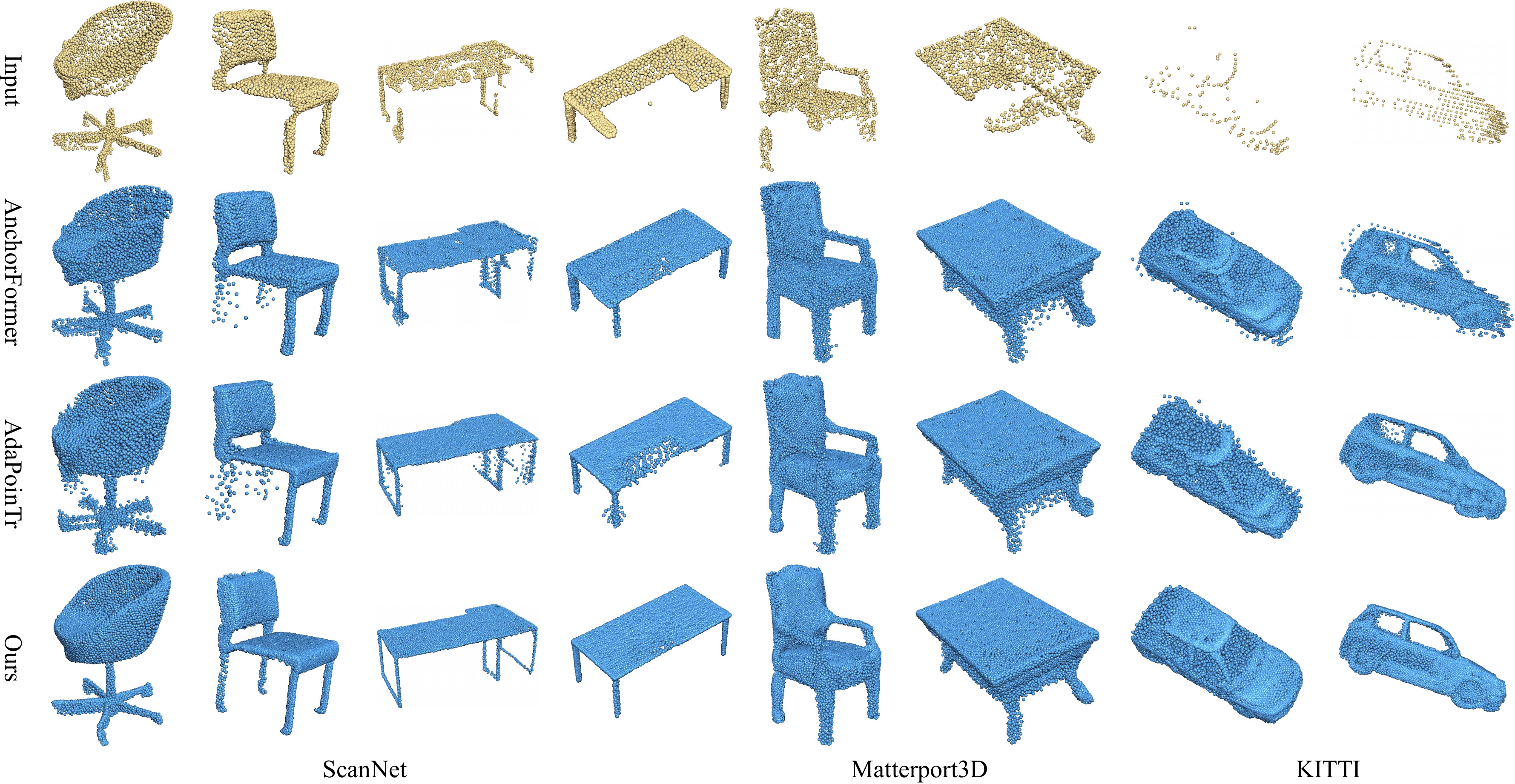}
\caption{Visual comparisons with recent methods~\citep{chen2023anchorformer,10232862} on real-world scans.}
  \label{fig:real}
\end{figure*}

\subsubsection{Quantitative Results}
We compare PointSea with 17 competitors~\citep{yuan2018pcn,xie2020grnet,wang2020cascaded,zhang2020detail,yu2021pointr,9928787,wen2022pmp,yan2022fbnet,zhou2022seedformer,zhang2022point,tang2022lake,fu2023vapcnet,DBLP:journals/ijcv/ZhangLXNZTL23,10097548,chen2023anchorformer,10232862,Zhu_2023_ICCV} in Tab.~\ref{tab:pcnl1}. The results demonstrate that PointSea achieves the best performance across all metrics. In particular, the average CD value of PointSea is 6.35, which is 0.19 lower than the CD value of the second-ranked method AdaPoinTr~\citep{10232862}. Meanwhile, our approach also achieves the lowest CD value among all 8 categories. 

\subsubsection{Qualitative Results}
In Fig.~\ref{fig:Vis_PCN}, we present a visual comparison of the results produced by different methods, revealing that PointSea consistently generates more accurate completions. Take the table model in the second row as an example, only AnchorFormer, SVDFormer, and PointSea successfully locate and complete the missing regions, while PointSea produces a more compact and faithful result. 
Besides, in the case of the couch model in the fourth row, a thin cross-section is demonstrated to further reveal the superiority of our method. Although all methods can generate the overall shape, PointSea excels at reconstructing the sharp edges of the handrail with fewer noisy points.

\subsection{Evaluation on Projected-ShapeNet-55/34}
\subsubsection{Data and Metrics}
The Projected-ShapeNet-55/34 dataset was introduced in \citep{10232862}. The primary difference between this dataset and ShapeNet-55/34 is that incomplete point clouds are generated by noised back-projecting from 16 viewpoints, making it more challenging compared with simple back-projecting in PCN~\citep{yuan2018pcn} or online cropping in ShapeNet-55/34~\citep{10232862}. We use the $\displaystyle \ell ^{1}$ version of CD and F-Score@1\% as evaluation metrics.
\subsubsection{Results}
Tab.~\ref{tab:projectshapenet} details the performance of different methods across all categories of Projected-ShapeNet-55 and seen/unseen categories of Projected-ShapeNet-34. Our PointSea achieves the best performance among all the previous methods.
In Fig.~\ref{fig:Vis_p55}, we visually compare PointSea with two representative methods~\citep{10232862,chen2023anchorformer}. The comparison clearly demonstrates that PointSea excels in capturing the overall shape from noisy input and effectively completing the missing regions.
Moreover, we observe that when the input point clouds exhibit extreme noise, the network must refine the undesired points within these regions. Notably, methods that predict the missing part~\citep{yu2021pointr,chen2023anchorformer} lack this capability, preserving the noise in the final prediction. While AdaPoinTr~\citep{10232862} can refine some of these points, PointSea surpasses it by generating cleaner and more detailed structures.

\begin{table*}
    \renewcommand\arraystretch{1.2}
    \small
    \centering
    \caption{Quantitative results on real-world scans. Results on KITTI and MatterPort3D are generated by models pre-trained on the PCN dataset, while results on ScanNet are produced by models pre-trained on the Projected-ShapeNet-55 dataset. (MMD $\times 10^3$)}
    \label{tab:realworld}
    \begin{tabular}{|c|c|c|c|c|c|}
    \hline
    Methods & KITTI & ScanNet Chairs & ScanNet Tables & Matterport3D Chairs & Matterport3D Tables \\
    \hline
              GRNet~\citep{xie2020grnet}   & 5.350 & 3.042 & 1.955 & 1.382 & 1.062 \\
              SeedFormer~\citep{zhou2022seedformer}   & 1.179 & 2.916 & 1.451 & 1.206 & 0.986 \\
              AnchorFormer~\citep{chen2023anchorformer} & 1.252 & 3.259 & 1.381 & 1.334 & 0.952 \\
              AdaPoinTr~\citep{10232862}   & 1.058  & 2.585 & 1.191 & 1.242 & 0.931  \\
              SVDFormer~\citep{Zhu_2023_ICCV}   & 0.967 & 2.828 & 1.170 & 1.238 & 0.965 \\
              Ours   & \textbf{0.933}  & \textbf{2.362} & \textbf{0.943} & \textbf{1.067} & \textbf{0.902} \\
    \hline
    \end{tabular}
\end{table*}

\begin{figure}[h]
  \centering
  \includegraphics[width=\linewidth]{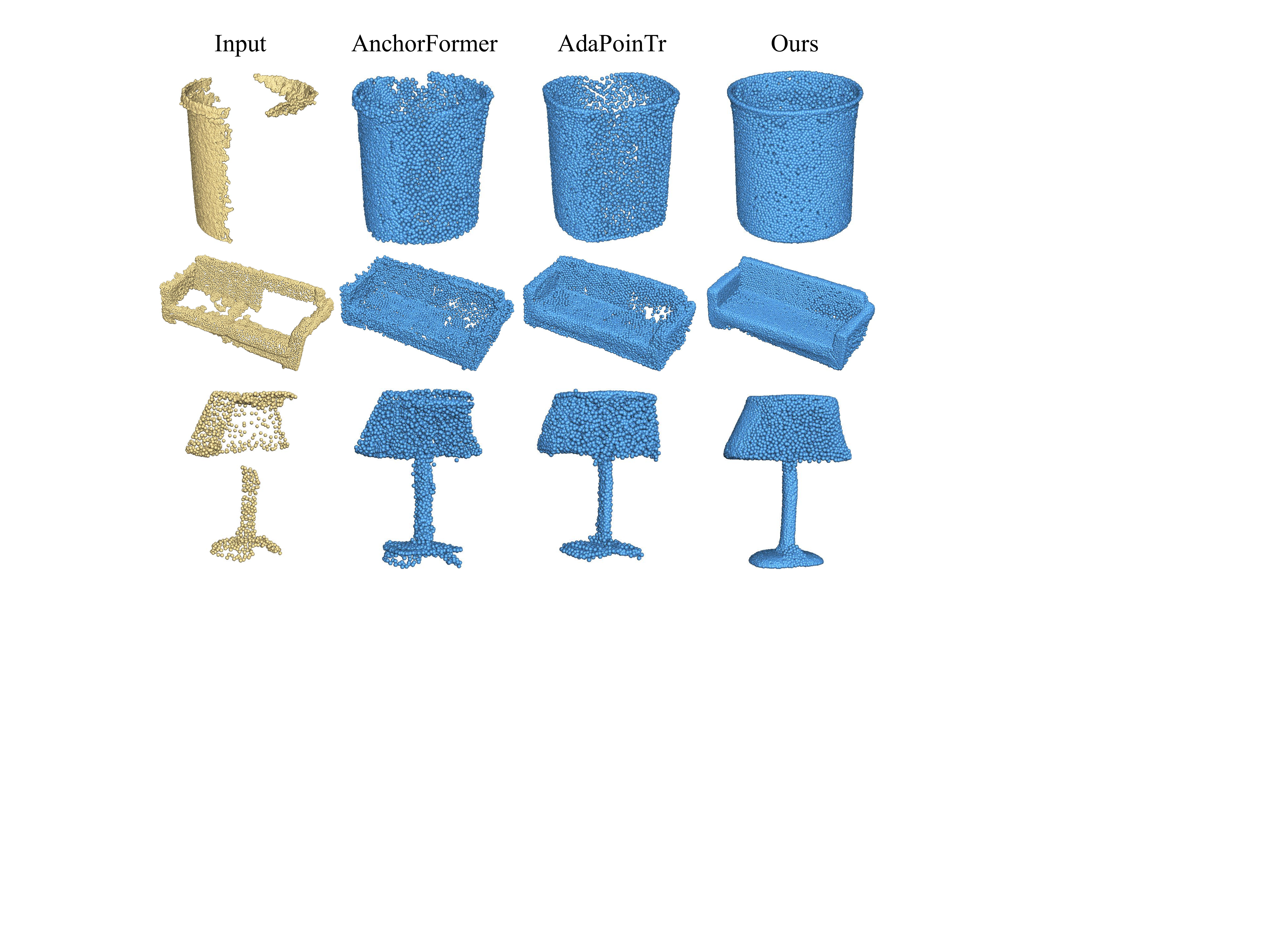}
\caption{Visual comparisons with recent methods~\citep{chen2023anchorformer,10232862} on more diverse real-world scans from \cite{dai2017scannet,choi2016large}.}
  \label{fig:diverse_real}
\end{figure}

\begin{table*}[h]
    \renewcommand\arraystretch{1.1}
    \centering
    \caption{Quantitative results on Semantic Scene Completion. ({$\displaystyle \ell ^{1}$} CD $\times 10^3$, mIoU, and mAcc)}
    \small
    \label{tab:ssc}
    \begin{tabular}{|c|ccc|ccc|}
    \hline
    \multirow{2}{*}{Methods} &  \multicolumn{3}{c|}{SSC-PC} & \multicolumn{3}{c|}{NYUCAD-PC} \\\cline{2-7} & CD$\downarrow$ & mIoU$\uparrow$ & mAcc$\uparrow$ & CD$\downarrow$ & mIoU$\uparrow$ & mAcc$\uparrow$ \\
    \hline
    Disp3D~\citep{wang2022learning} & 35.9 & 18.9 & 23.5 & 22.75 & 16.4 & 22.9 \\
    CasFusionNet~\citep{xu2023casfusionnet} & \textbf{8.96} & 91.3 & 94.8 & 10.28 & 49.5 & 59.7 \\
    \hline
    Ours &  9.17 & \textbf{92.2} & \textbf{95.3} & \textbf{9.52} & \textbf{54.6} & \textbf{65.4} \\
    \hline
    \end{tabular}
\end{table*}

\begin{figure*}[h]
  \centering
  \includegraphics[width=0.9\textwidth]{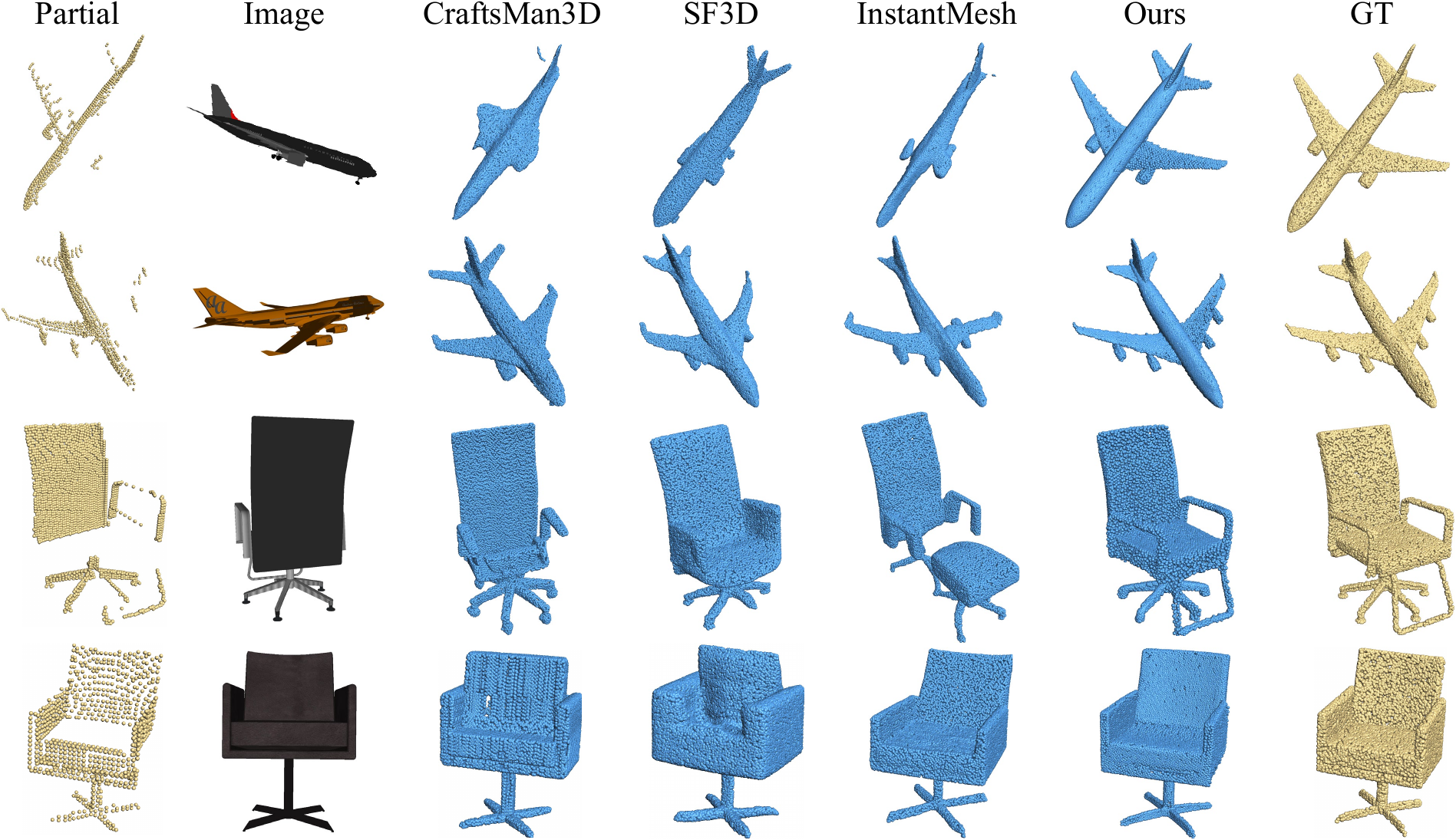}
\caption{Visual comparison with state-of-the-art image-to-3D models~\citep{li2024craftsman,sf3d2024,xu2024instantmesh} on the PCN dataset. To ensure a fair comparison, we render ShapeNet meshes from the same viewpoints used in the virtual scanning process of the PCN dataset and use the resulting images as inputs for these models.}
  \label{fig:gen_comp}
\end{figure*}

\begin{figure*}[h]
  \centering
  \includegraphics[width=0.95\textwidth]{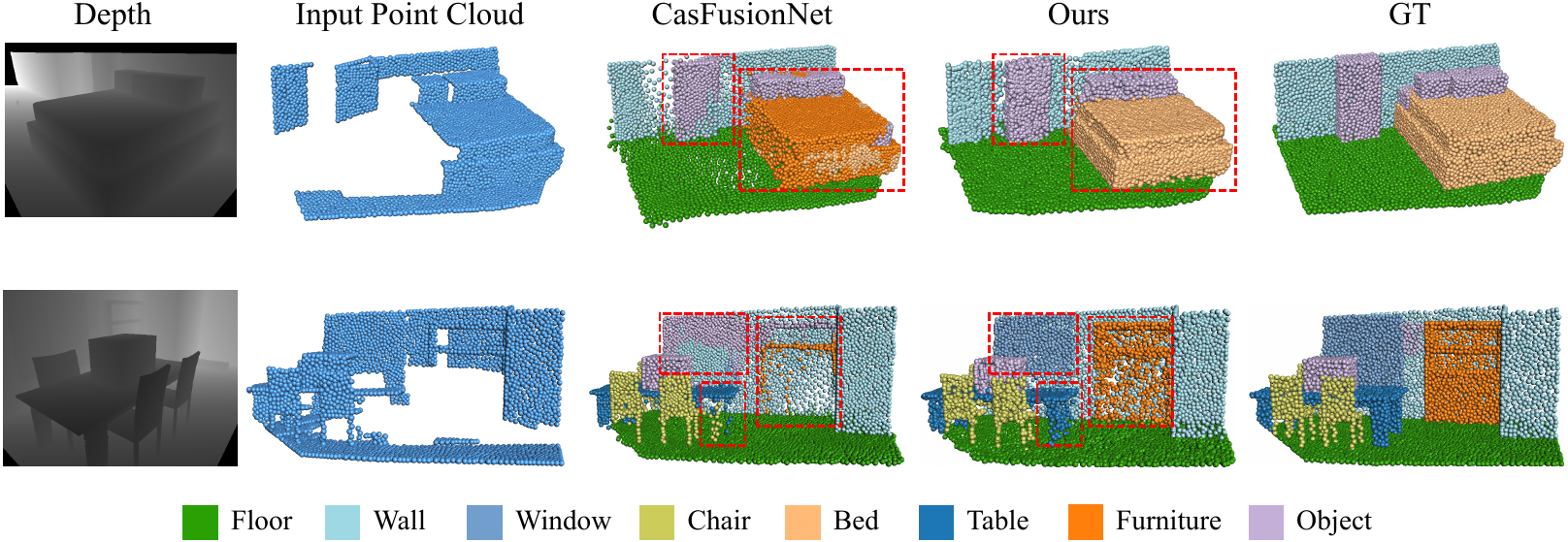}
\caption{Visual comparison with \citep{xu2023casfusionnet} on the NYUCAD-PC dataset. The original depth scans are shown in the leftmost column.}
  \label{fig:ssc}
\end{figure*}

\subsection{Evaluation on Real-world Scans}
\subsubsection{Data and Metrics}
To evaluate our method on real-world scans captured by different sensors, we conducted tests on a diverse set of point clouds: 2,401 car point clouds from KITTI~\citep{geiger2013vision}, 100 chair and table point clouds from ScanNet~\citep{dai2017scannet}, and 20 chair and table point clouds from Matterport3D~\citep{matterport3d}.
Results on KITTI and MatterPort3D are generated by models pre-trained on the PCN dataset,
while results on ScanNet are produced by models pre-trained on the Projected-ShapeNet-55 dataset.
Since there is no ground truth available for real-world scans, our quantitative evaluation metric is the Minimal Matching Distance (MMD) calculated based on the $\displaystyle \ell ^{2}$ version of CD. 
In this context, MMD represents the CD value between the output point cloud and the most similar car/table/chair point cloud from ShapeNet, based on CD. This metric quantifies how closely the output resembles a typical car, table, or chair.

\subsubsection{Results}
The quantitative results are reported in Tab.~\ref{tab:realworld}. The superior performance in terms of MMD indicates that results produced by PointSea closely resemble those of a typical car/chair/table, even when the input data exhibits a distribution shift from the training set.
Additionally, we demonstrate the visual comparisons with two representative methods~\citep{chen2023anchorformer,10232862} in Fig.~\ref{fig:real}. 
To provide a more comprehensive evaluation, we also show results on more diverse real-world scans from ScanNet~\citep{dai2017scannet} and Redwood~\citep{choi2016large} in Fig~\ref{fig:diverse_real}, which not only expands the category diversity but also introduces more irregular types of missing regions.
It is evident that, in scenarios where the point clouds captured by real-world sensors contain noisy points or exhibit inconsistent scale compared to the training data, our method demonstrates superior performance in reconstructing a smooth surface and preserving the original structures.

\subsection{Comparison with Image-to-3D Models}
Given the success of emerging 3D generative models~\citep{wang2025crm,tang2024dreamgaussian}, we compare PointSea with three state-of-the-art image-to-3D models~\citep{li2024craftsman,sf3d2024,xu2024instantmesh}, as shown in Fig.~\ref{fig:gen_comp}. 
To fairly evaluate the potential of image-to-3D models, we first render ShapeNet meshes from the same viewpoints used in the virtual scanning process of the PCN dataset. The resulting images serve as inputs for these models.
The results in the first, second, and third rows reveal that when input images are affected by self-occlusion, generative models struggle to reconstruct unseen parts, often resulting in distorted or undesired shapes. Additionally, without leveraging geometric cues like point clouds, these generative models fail to maintain the original structure of the partial input. For instance, in the fourth row, although all generative methods produce visually plausible chairs, they fail to reconstruct the correct back angle due to the limited information provided by the input images.
In contrast, our SDG's dual-way design ensures both accurate reconstruction of unseen parts and faithful preservation of the input geometry.

\subsection{Extension to Semantic Scene Completion}

\subsubsection{Network Adjustment}
In this experiment, our objective is to assess the performance of PointSea when dealing with complex scene-level point clouds and to explore its potential application in semantic understanding.
Given a partial observation of a scene, the goal of semantic scene completion methods is to reconstruct the completion scene along with the semantic category of each point.
The key difference between this task and object completion is the additional prediction of semantic labels.
To adjust PointSea to this task, we unleash the full potential of SDG to simultaneously predict the geometric structure and semantic category in a coarse-to-fine manner.
Specifically, each SDG is equipped with an MLP-based segmentation head. In the first SDG, the offset feature $F_{1}$ is fed into the attached head to produce the probability of each class $\emph{L}_{1}\subseteq\mathbb{R}^{N_{1} \times C_N}$ for $P_1$, where $C_N$ represents the number of classes. In the subsequent SDG, the segmentation results are predicted through residual learning. In particular, $F_{2}$ is input into the attached head to regress a probability offset $\emph{L}_{2}$. Then, the predicted probability of this step is obtained by adding $\emph{L}_{2}$ to $\emph{L}_{1}$.
When the point location shifts across different class regions during refinement, this design ensures that the semantic label can adaptively change along with it since $\emph{L}_{l}$ and $O_l$ share the same information.
Furthermore, a modification is introduced in the loss function. In addition to the completion loss in Eq.~\ref{eqnloss}, we need another loss function to supervise the semantic segmentation. Since there is no direct one-to-one mapping between $\emph{L}_{l}$ and the ground-truth labels, we first establish the mapping when calculating CD and leverage this mapping to find the corresponding label for each point. After that, cross-entropy loss is computed to supervise the segmentation results in each SDG. 

\subsubsection{Dataset and Metrics}
We conduct experiments on the SSC-PC and NYUCAD-PC datasets, as provided in \cite{xu2023casfusionnet}, to evaluate the performance of PointSea on SSC. SSC-PC is created based on the dataset provided in \cite{zhang2021point}. The input incomplete point clouds are generated through virtual scanning, with both the input and output point clouds comprising 4,096 points. NYUCAD-PC is created based on the NYUCAD dataset~\citep{Firman_2016_CVPR}. The input incomplete point clouds, consisting of 4,096 points, are generated through real-world depth scans, while the ground-truth point clouds, consisting of 8,192 points, are sampled from CAD mesh annotations. We follow the same experimental setting with \cite{xu2023casfusionnet}.
We use $\displaystyle \ell ^{1}$ version of CD as the evaluation metric of completion quality. To evaluate the semantic segmentation results, we adopt the mean class IoU (mIoU) and the mean class accuracy (mAcc) as metrics.
\subsubsection{Results}
We compare our method with two state-of-the-art methods: Disp3D~\citep{wang2022learning} and CasFusionNet~\citep{xu2023casfusionnet}. The quantitative results are given in Tab.~\ref{tab:ssc}, from which we can find that PointSea achieves the best performance on the real-world dataset NYUCAD-PC. On the SSC-PC dataset, PointSea ranks second in terms of CD but outperforms in the segmentation metrics.
In Fig.~\ref{fig:ssc}, we further illustrate the superior performance of PointSea by visually comparing our method with the state-of-the-art CasFusionNet~\citep{xu2023casfusionnet}. Clearly, the complete scenes generated by PointSea exhibit finer local structures and yield more accurate segmentation results.
Overall, the results demonstrate the effectiveness of our SDG in handling complex scenes and highlight its potential for extension to other tasks.

\subsection{Ablation Studies}
\label{ablationSec}
To ablate PointSea, we remove and modify the main components. All ablation variants are tested on the PCN and ShapeNet-55 datasets. The ablation variants can be categorized as ablations on SVFNet and SDG.

\begin{table*}
        \renewcommand\arraystretch{1.2}
        \centering
        \caption{
        Effect of SVFNet. (PCN: {$\displaystyle \ell ^{1}$} CD $\times 10^3$ and F1-Score@1\%. ShapeNet-55: {$\displaystyle \ell ^{2}$} CD $\times 10^4$ and F1-Score@1\%)
        }
        \label{tab:ablationSVFNet}
        \footnotesize
        \begin{tabular}{|c|c c c c c| c c | c c|}
        \hline
        \multirow{2}{*}{Methods}  & \multirow{2}{*}{Sparse Projection} & \multirow{2}{*}{Dense Projection} & \multirow{2}{*}{Intra-view} & \multirow{2}{*}{Inter-view} & \multirow{2}{*}{One-stage} & \multicolumn{2}{c|}{PCN} & \multicolumn{2}{c|}{ShapeNet-55} 
        \\ \cline{7-10} & &&&& & CD$\downarrow$ & F1$\uparrow$ & CD$\downarrow$ & F1$\uparrow$ \\ 
        \hline
        A : w/o Projection  & & & & & & 6.53  & 0.836 & 8.46 & 0.451\\
        B : w/o Feature Fusion & & \checkmark & & & & 6.58 & 0.831 & 8.52 & 0.445\\
        C : Intra-view Fusion & & \checkmark & \checkmark & & & 6.43 & 0.847 & 7.93 & 0.476\\
        D : Inter-view Fusion & & \checkmark & & \checkmark & & 6.49 & 0.845 & 8.23 & 0.458 \\
        E : One-stage Fusion & & \checkmark & & & \checkmark & 6.46 & 0.848 & 7.88 & 0.469\\
        F : Sparse Projection & \checkmark & & \checkmark & \checkmark & & 6.39 & 0.853 & 7.80 & 0.480\\
        Ours  & & \checkmark & \checkmark & \checkmark & & \textbf{6.35} & \textbf{0.858} & \textbf{7.53} & \textbf{0.485}  \\
        \hline
        \end{tabular}
\end{table*}

\begin{table}
        \renewcommand\arraystretch{1.2}
        \centering
        \caption{Effect of projection numbers on the PCN dataset. ({$\displaystyle \ell ^{1}$} CD $\times 10^3$, DCD, F1-Score@1\%, and Inference Time)}
        \label{tab:ablationViewNumber}
        \small
        \begin{tabular}{|c|c c c |c|}
        \hline
        Methods  & CD$\downarrow$ & DCD$\downarrow$ & F1$\uparrow$ & Time\\
        \hline
        1 view  &  6.44 & 0.528 & 0.848  & 23.51ms\\
        3 views (Ours) & \textbf{6.35} & \textbf{0.522} & 0.858 & 24.92ms\\
        6 views & 6.37 & 0.524 & \textbf{0.860} & 26.77ms\\
        \hline
        \end{tabular}
\end{table}

\begin{table}
        \renewcommand\arraystretch{1.2}
        \centering
        \caption{Robustness to random perturbation of projection on the PCN dataset. ({$\displaystyle \ell ^{1}$} CD $\times 10^3$, DCD, and F1-Score@1\%)}
        \label{tab:ablationNoise}
        \footnotesize
        \resizebox{0.4\textwidth}{!}{
        \begin{tabular}{|cc|ccc|}
        \hline
        \multicolumn{2}{|c|}{Perturbation Levels} & \multirow{2}{*}{CD$\downarrow$} & \multirow{2}{*}{DCD$\downarrow$} & \multirow{2}{*}{F1$\uparrow$} \\ \cline{1-2} Distance & Angle &&& \\
        \hline
        \multicolumn{2}{|c|}{w/o Perturbation}&  \textbf{6.35} & \textbf{0.522} & \textbf{0.858} \\
        $0.05$ & $5^{\circ}$  &  6.36 & 0.523 & 0.857 \\
        $0.1$ & $10^{\circ}$ & 6.38 & 0.524  & 0.856 \\
        $0.15$ & $15^{\circ}$ & 6.43 & 0.526  & 0.853 \\
        $0.2$ & $20^{\circ}$ & 6.51 & 0.539 & 0.849\\
        \hline
        \end{tabular}
        }
\end{table}

\begin{table}[t]
        \renewcommand\arraystretch{1.2}
        \centering
        \caption{Effect of different 2D encoders on the PCN dataset. ``ImageNet'' and ``DINO V2'' denote models pre-trained by ImageNet~\citep{5206848} classification and DINO V2~\citep{dinov2}. ({$\displaystyle \ell ^{1}$} CD $\times 10^3$, DCD, F1-Score@1\%, and Inference Time)}
        \label{tab:ablation2Dbackbone}
        \small
        \begin{tabular}{|c|c c c|c|}
        \hline
        Methods  & CD$\downarrow$ & DCD$\downarrow$ & F1$\uparrow$ & Time \\
        \hline
        ResNet-18 (w/o Pre-trained) & 6.38 & 0.524 & 0.854 & 24.92ms \\
        ResNet-18 (ImageNet, Ours) & 6.35 & 0.522 & 0.858 & 24.92ms\\
        Resnet-50 (w/o Pre-trained) & 6.37 & 0.524 & 0.856 & 27.53ms\\
        Resnet-50 (ImageNet) & 6.33 & 0.520 & 0.861 & 27.53ms\\
        Vit-B/16 (w/o Pre-trained) & 6.43 & 0.527 & 0.851 & 42.81ms\\
        Vit-B/16 (ImageNet) & 6.38 & 0.525 & 0.850 & 42.81ms\\
        Vit-B/14 (w/o Pre-trained) & 6.44 & 0.525 & 0.853 & 46.52ms\\
        Vit-B/14 (DINO V2) & \textbf{6.29} & \textbf{0.518} & \textbf{0.867} & 46.52ms\\
        \hline
        \end{tabular}
\end{table}

\begin{table*}[t]
        \renewcommand\arraystretch{1.2}
        \centering
        \caption{
        Effect of SDG. (PCN: {$\displaystyle \ell ^{1}$} CD $\times 10^3$ and F1-Score@1\%. ShapeNet-55: {$\displaystyle \ell ^{2}$} CD $\times 10^4$ and F1-Score@1\%)
        }
        \label{tab:ablationIDTr}
        \footnotesize
        \begin{tabular}{|c|c c c c|c c | c c|}
        \hline
        \multirow{2}{*}{Methods}  & \multirow{2}{*}{Analysis} & \multirow{2}{*}{Alignment} & \multirow{2}{*}{Embedding} & \multirow{2}{*}{Selection} & \multicolumn{2}{c|}{PCN} & \multicolumn{2}{c|}{ShapeNet-55} 
        \\ \cline{6-9} & &&& & CD$\downarrow$ & F1$\uparrow$ & CD$\downarrow$ & F1$\uparrow$ \\
        \hline
        G : w/o Embedding & \checkmark & \checkmark & & & 6.56  & 0.839 & 8.16 & 0.454 \\
        H : w/o Alignment& \checkmark & & \checkmark & & 6.69  & 0.828 & 8.83 & 0.437\\
        I : w/o Analysis & &\checkmark &  & &  6.71   & 0.825 & 9.08 & 0.412\\ 
        J : w/o Selection &\checkmark&\checkmark &\checkmark & & 6.46  & 0.845 & 7.93 & 0.466\\
        Ours & \checkmark & \checkmark & \checkmark & \checkmark & \textbf{6.35} &\textbf{0.858} & \textbf{7.53} & \textbf{0.485}  \\
        \hline
        \end{tabular}
\end{table*}

\begin{table}
        \renewcommand\arraystretch{1.2}
        \centering
        \caption{Effect of the number of SDG modules on the PCN dataset. ({$\displaystyle \ell ^{1}$} CD $\times 10^3$, DCD, F1-Score@1\%, and Inference Time)}
        \label{tab:ablationSDGNumber}
        \small
        \begin{tabular}{|c|c c c |c|}
        \hline
        Methods  & CD$\downarrow$ & DCD$\downarrow$ & F1$\uparrow$ & Time\\
        \hline
        1 SDG  &  6.56 & 0.549 & 0.841  & 15.86ms\\
        2 SDGs (Ours) & \textbf{6.35} & 0.522 & \textbf{0.858} & 24.92ms\\
        3 SDGs & 6.36 & \textbf{0.520} & \textbf{0.858} & 33.53ms\\
        \hline
        \end{tabular}
\end{table}

\subsubsection{Ablation on SVFNet}
To investigate the impact of shape descriptor extraction methods, we compare six variants of SVFNet, and the results are presented in Tab.~\ref{tab:ablationSVFNet}. 
In variant A, we remove the input depth maps, and the completion performance is limited by relying only on 3D coordinates to understand shapes. 
In variant B, we evaluate the importance of our Feature Fusion module by replacing the fusion of different inputs with late fusion, which directly concatenates features from different modalities. We observe an evident drop in performance, indicating that the proposed SVFNet can effectively fuse cross-modal features. 
Then, we evaluate the effect of different feature fusion strategies. In variant C, same with SVDFormer~\citep{Zhu_2023_ICCV}, fusion happens only at the intra-view level. In variant D, conversely, we only use inter-view feature fusion. Whether employing intra-view or inter-view fusion individually, the performance is inferior compared to their combination. In variant E, instead of fusing features in two stages, we reshape patched features $F_V$ from all three views to the size of ${N_V H^{\prime} W^{\prime} \times C}$ and feed it to a single fusion module. The decreased performance indicates that fusing patch-wise features from all views weakens the model's ability to perceive shapes from multiple perspectives.
Finally, in variant F, we use the vanilla perspective projection applied in SVDFormer~\citep{Zhu_2023_ICCV} to replace the dense projection~\citep{PointCLIPV2}. We see the performance of our method slightly drops due to the insufficient semantic information contained in the sparse depth maps. 

\begin{figure}[h]
  \centering
  \includegraphics[width=\linewidth]{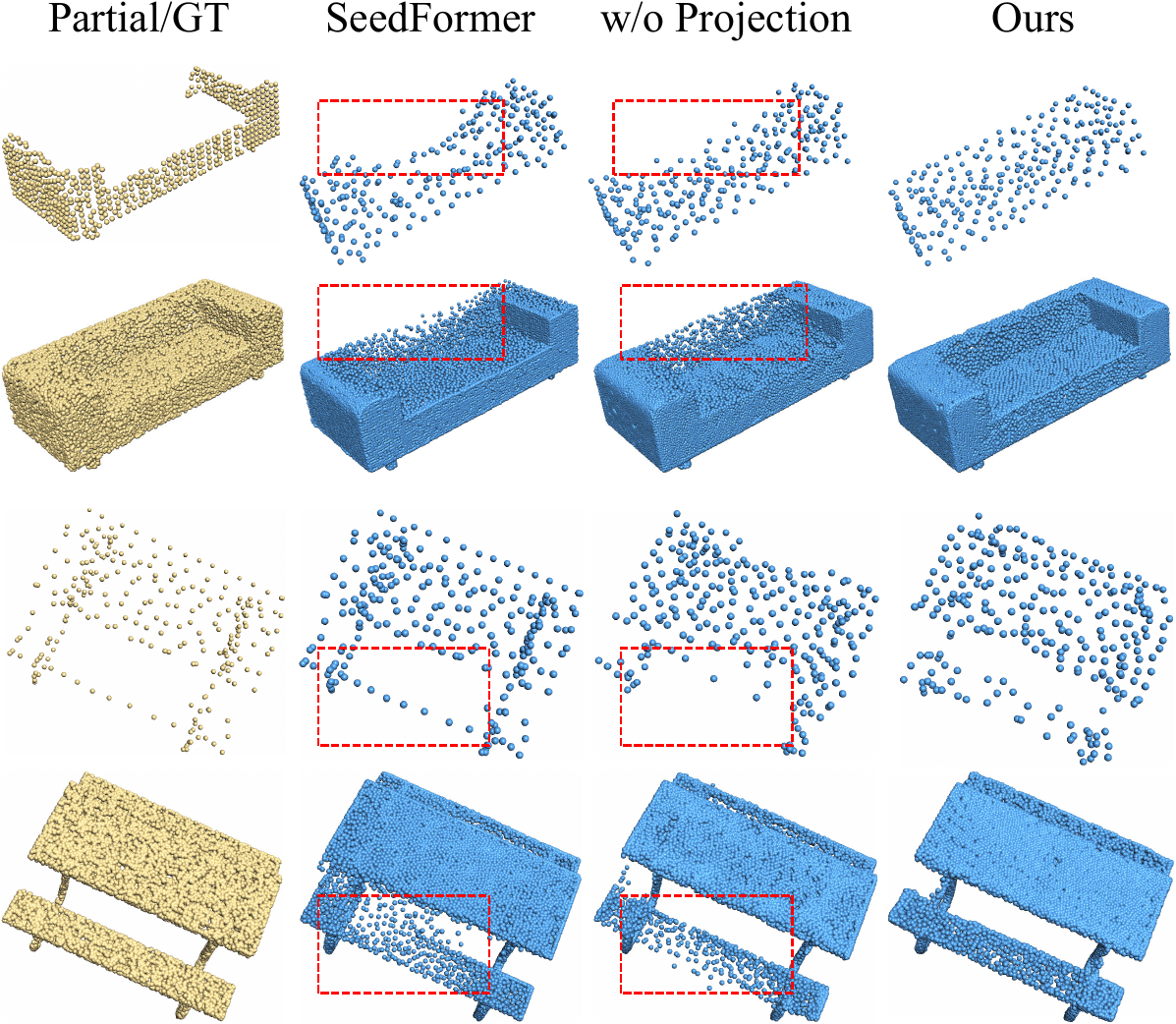}
\caption{Visual comparison of the representative coarse-to-fine method \citep{zhou2022seedformer}, variant A (w/o projection), and our method on two partial models. The upper results are generated coarse results.}
  \label{fig:MVFVis}
\end{figure}

Furthermore, to conduct a more thorough analysis of the effectiveness of our SVFNet, we visualize the results produced by our approach, variant A, and SeedFormer, which also employ the coarse-to-fine paradigm. In Fig.~\ref{fig:MVFVis}, we present the results alongside the coarse point cloud generated directly by SVFNet (patch seeds of \cite{zhou2022seedformer}). Our analysis reveals that during the initial completion stage, both SeedFormer and variant A produce suboptimal results, such as generating too few points in missing areas. This presents a challenge for the refinement stage, making it difficult to produce satisfactory final results. Our SVFNet overcomes this challenge by leveraging multiple viewpoints to observe partial shapes. By doing so, our method can locate missing areas and generate a compact global shape, leading to the production of fine details in the results. 

\begin{figure}[h]
  \centering
  \includegraphics[width=0.97\linewidth]{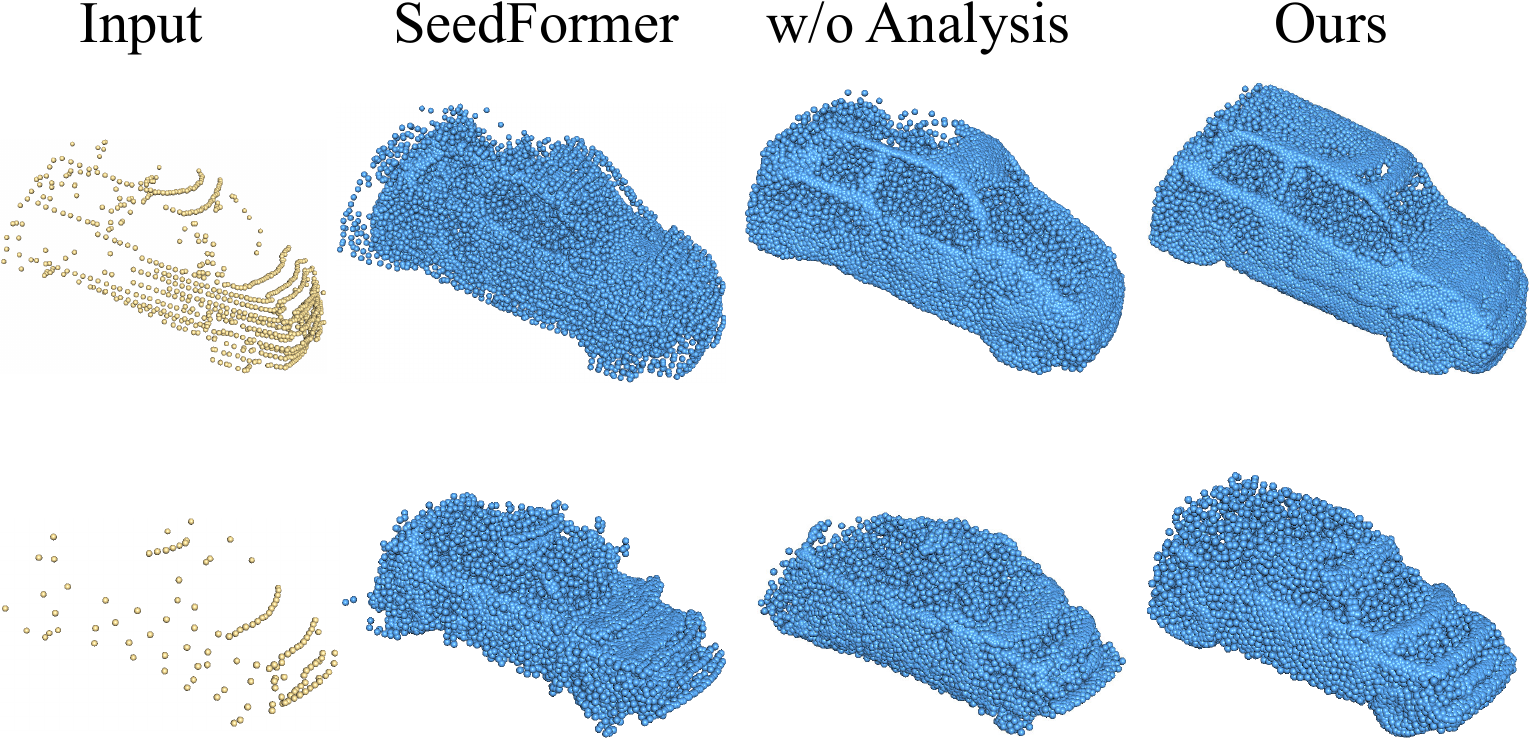}
\caption{Visual comparison with SeedFormer~\citep{zhou2022seedformer} and variant H (w/o Structure Analysis) on LiDAR scans from KITTI.}
  \label{fig:Ablation_Analysis}
\end{figure}

\begin{figure*}[h]
  \centering
  \includegraphics[width=\textwidth]{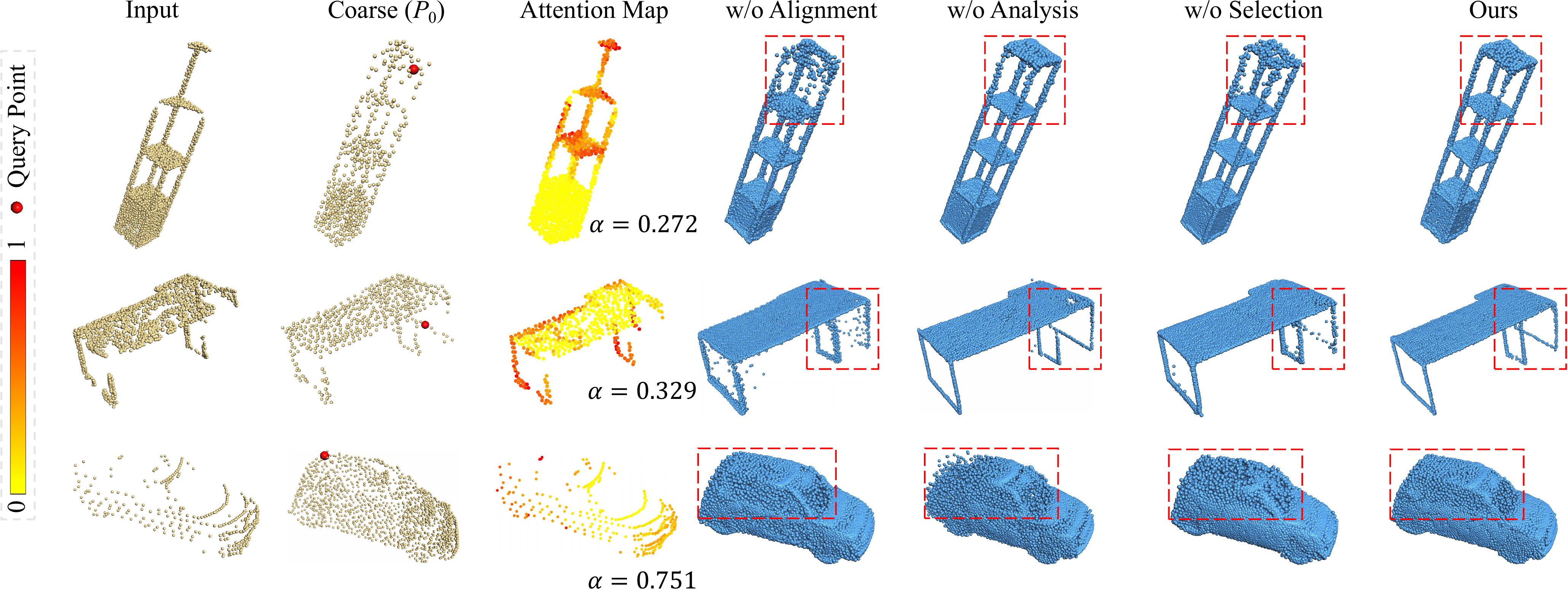}
\caption{
Visual comparison of generated coarse point cloud $P_0$ and results of variant H (w/o Similarity Alignment), I (w/o Structure Analysis), and J (w/o Selection). 
We select a query point (marked with red) in $P_0$ and visualize the attention map in the cross-attention layer. The redder the color, the higher the similarity. The mean $\alpha$ values of the selected point from the Path Selection module are also listed in the rightmost column.
}
  \label{fig:Ablation_cross}
\end{figure*}

\noindent\textbf{Effect of the number of projections.} 
We conduct an ablation experiment to assess the impact of varying the number of projections. The depth maps are projected from one front view, as well as from three and six orthogonal views, respectively. The results, presented in Tab.~\ref{tab:ablationViewNumber}, reveal a challenge for the network in perceiving the whole shape when using the information from only a single view.
Simultaneously, we see that the performance will not continue increasing when utilizing all six views.
This can be attributed to the following reasons:
Traditional methods, such as multi-view stereo, rely on extensive correspondences across dozens of RGB images and geometric constraints from well-calibrated camera poses to infer 3D shapes. As a result, incorporating more views typically leads to improved 3D reconstruction quality.
In contrast, our approach utilizes multi-view depth images projected from partial point clouds. These depth maps provide only rough structural information and lack texture details, making them less informative compared to RGB images. 
Given this limitation, we employ only a few views and implicitly fuse them with point clouds and viewpoint information in the feature level to predict the coarse shape. The results indicate that three views are sufficient for this task. Increasing the number of views tends to introduce information overlap and redundancy, and may also introduce errors.

\noindent\textbf{Robustness to random perturbation of projection.} 
To assess the robustness of our method, during inference, we introduce zero-mean Gaussian noise to the projection process. We add perturbation to the projection by including camera view angle offsets and observation distance displacements. We varied the perturbation levels by adjusting the standard deviations to four different values (0.05, 0.1, 0.15, 0.2 unit length for observing distance and 5, 10, 15, 20 degrees for camera angles). The results in Tab.~\ref{tab:ablationNoise} demonstrate that PointSea is robust to slight perturbations. As the noise level increases significantly ({$0.2$, $20^{\circ}$}), there is a noticeable drop in performance. This can be attributed to the substantial noise, which causes the information contained among the three projections to become partially overlapped and lost. Consequently, this poses a challenge for the feature fusion.

\noindent\textbf{Effect of different 2D backbones.} 
We ablate the choice of 2D backbone in the SVFNet. To be specific, we replace it with ResNet-50~\citep{he2016deep} and the vision transformer~\citep{dosovitskiy2021an}, respectively. The results in Tab.~\ref{tab:ablation2Dbackbone} indicate that a larger 2D backbone does not yield a substantial performance improvement, but it does introduce increased computational costs. 
Moreover, we ablate the pre-trained weights of 2D encoders. We first use the commonly used ImageNet~\citep{5206848} pre-training and find that models pre-trained in this way perform slightly better than those trained from scratch.
Then, to explore the potential of vision foundation models in the completion task, we use the recently proposed DINO V2 model~\citep{dinov2} in PointSea. The results demonstrate that visual features derived from large-scale self-supervised learning can significantly enhance completion performance, compared with pre-training on ImageNet or training from scratch.
However, the ViT model introduces a substantial computational burden, increasing the parameter number and inference time by more than twofold compared to the default setting. For this reason, we continue to use ResNet-18 pre-trained on ImageNet as the 2D encoder in our main experiments.

\subsubsection{Ablation on SDG}
Tab.~\ref{tab:ablationIDTr} compares different variants of PointSea on the SDG module. In variant G, we remove the Incompleteness Embedding of SDG, which results in a higher CD value and a lower F1-Score, indicating that the ability to perceive the incompleteness degree of each part is crucial for the model. In the variants H and I, we completely remove the Similarity Alignment and Structure Analysis path from SDG, respectively. The results show that the performance of the model significantly drops when any one of these paths is removed.
Compared with Variant J, where only the path selection module is removed, the model can flexibly choose features from different paths and achieve the best performance.

\noindent\textbf{Effect of the number of SDG modules.} 
We conducted an ablation experiment to evaluate the impact of varying the number of SDG modules. In our main experiments, we stack two SDGs with upsampling rates of \{4, 8\}. For comparison, we now evaluate configurations with a single SDG using an upsampling rate of \{32\} and three SDGs with upsampling rates of \{2, 2, 8\}. The results are presented in Tab.~\ref{tab:ablationSDGNumber}.
We find that directly upsampling point clouds by a factor of 32 results in a significant performance drop.
This indicates that the refinement process relies on the collaboration of multiple SDGs, which regards shape refinement as a cascaded process. 
Then, when the number of SDGs is increased to three, the DCD value shows slight improvement, but the CD value remains unchanged, indicating that two SDGs are adequate for generating 16,384 points in the PCN dataset.
Note also that using more SDGs may become necessary when generating a higher number of points.

\noindent\textbf{Visual comparison of different variants} 
To better understand SDG, we show more visual results in Fig.~\ref{fig:Ablation_Analysis} and Fig.~\ref{fig:Ablation_cross}. 
Specifically, we investigate the effectiveness of the two units and the Path Selection module by comparing the performance of different variants of the model on different kinds of incomplete point clouds. 
In Fig.~\ref{fig:Ablation_Analysis}, our method can generate plausible shapes, while the variant I and SeedFormer produce undesired structures due to over-preserving the partial input. This result proves the importance of the Structure Analysis path, particularly when the input contains limited information. 
Fig.~\ref{fig:Ablation_cross} shows the results produced by different variants H, I, J, and ours.
We visualize the attention map in the cross-attention layer to demonstrate the effectiveness of the Similarity Alignment unit. 
The query point (marked in red) is absent in the input and shown in the coarse result $P_0$. 
The mean $\alpha$ (in Eq.~\ref{eqn3}) value of the selected query point is also provided.
The results demonstrate that for shapes with highly similar regions (lamp in the first row and table in the second row), the Similarity Alignment unit effectively identifies similar geometries over short or long distances, resulting in finer details. The model can produce plausible shapes with fine details even without the Structure Analysis Path. However, without the Path Selection module, the effectiveness of the Similarity Alignment unit is limited. In such cases, the Path Selection module assigns smaller 
$\alpha$ values, prioritizing the Alignment unit and leading to improved results.
For shapes characterized by highly sparse and noisy partial input (LiDAR scan in the third row), the attention map shows that the Alignment unit struggles to extract useful information from similar regions. Here, a larger 
$\alpha$ value directs the model to focus more on the Analysis unit, mitigating the influence of noisy information in the partial input.
Based on these observations, we conclude that the Path Selection module dynamically aggregates diverse features, thereby enhancing detail completion.

\begin{table}[h]
        \renewcommand\arraystretch{1.2}
        \centering
        \caption{Complexity analysis. We report the inference time (ms) and the number of parameters (Params) on the PCN dataset (16,384 points). Our method achieves a balance between computation cost and performance.}
        \label{tab:complexity}
        \footnotesize
        \begin{tabular}{|c|cc|c|} 
        \hline
        Methods  & \makebox[0.02\linewidth][c]{Time} & \makebox[0.02\linewidth][c]{Params} & CD$\downarrow$\\
        \hline
        GRNet~\citep{xie2020grnet}  & 14.51ms & 76.71M & 8.83    \\
        SeedFormer~\citep{zhou2022seedformer}  & 43.09ms & 3.24M & 6.74    \\
        GTNet~\citep{DBLP:journals/ijcv/ZhangLXNZTL23} & 62.53ms & 13.48M & 7.15    \\
        AdaPoinTr~\citep{yu2021pointr}  & 15.26ms & 32.49M & 6.53    \\
        SVDFormer~\citep{Zhu_2023_ICCV}  & 26.55ms & 32.63M & 6.54    \\
        \hline
        PointSea  & 24.92ms & 41.22M & 6.35  \\
        PointSea-L  & 16.52ms & 20.55M & 6.47 \\
        \hline
        \end{tabular}
\end{table}

\subsubsection{Complexity Analysis}
We present the complexity analysis in Tab.~\ref{tab:complexity}, including the inference time on a single NVIDIA 3090 GPU, the number of parameters, and the corresponding results on the PCN dataset. The notable increase in model size compared to SVDFormer is primarily attributed to the incorporation of a complete ResNet-18 to harness the capabilities of pre-trained 2D models.
To strike a favorable balance between cost and performance, we introduce a more lightweight version of PointSea, denoted as PointSea-L in Tab.~\ref{tab:complexity}. PointSea-L reduces the number of hidden dimensions in attention layers by half and, akin to SVDFormer, employs a compact ResNet-18 with 1/4 the feature size of the original ResNet. The results indicate that PointSea can achieve state-of-the-art performance with faster inference speed and fewer parameters compared to its previous version.

\label{secEXP}

\begin{figure}[h]
  \centering
  \includegraphics[width=\linewidth]{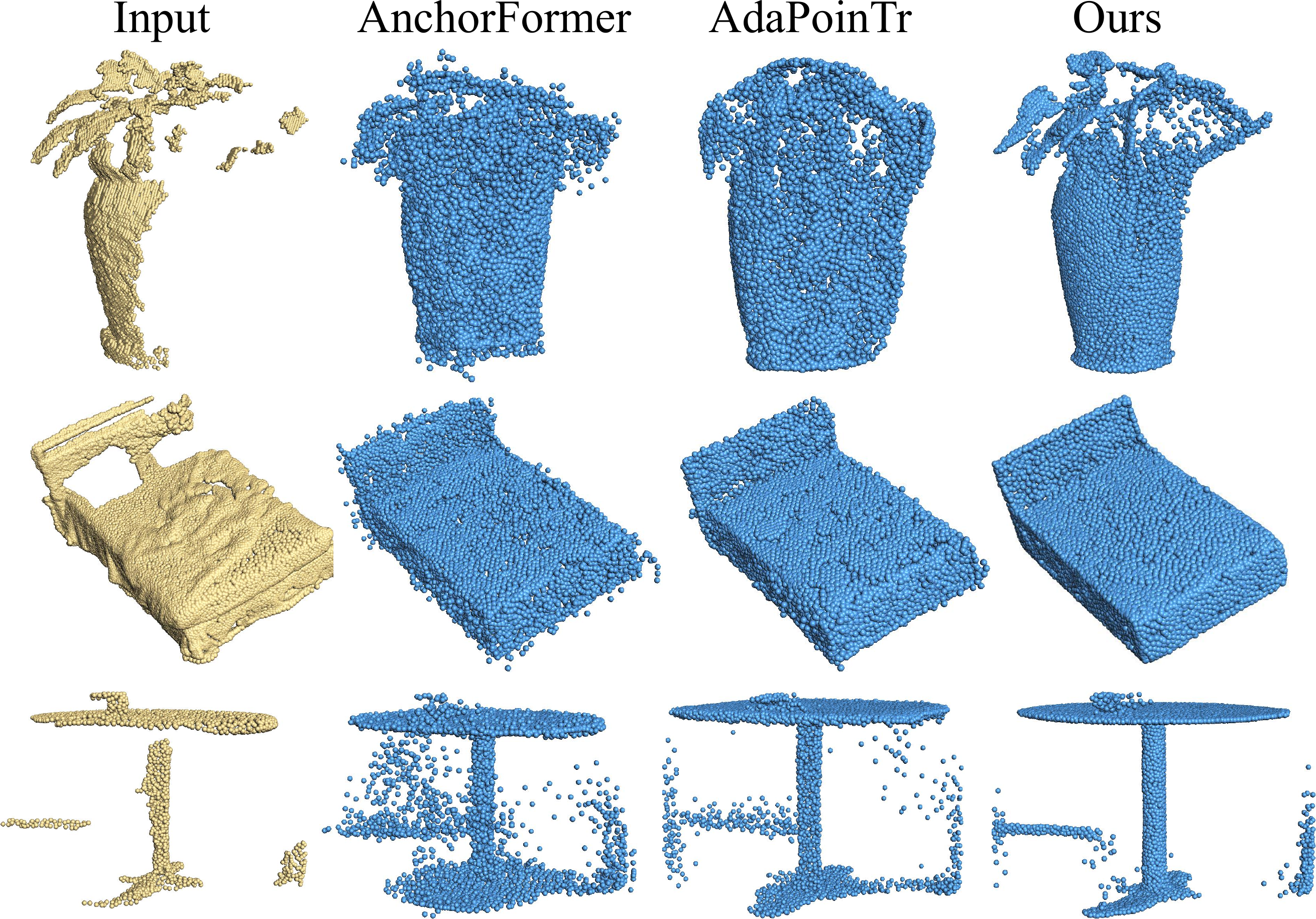}
\caption{
Visualization of failure cases. PointSea may generate inaccurate local details when the input point clouds exhibit substantial domain gaps from the training data, such as highly complex topologies or outliers from other objects. Nevertheless, even in these challenging scenarios, PointSea outperforms competitors~\citep{chen2023anchorformer,10232862} by reconstructing more plausible shapes.
}
  \label{fig:cases}
\end{figure}

\section{Conclusion}
We propose PointSea for point cloud completion. We start by identifying the main challenges in the completion and developing new solutions for each of them. PointSea leverages self-projected multi-view analysis to comprehend the overall shape and effectively perceive missing regions. Furthermore, we introduce a decoder called Self-structure Dual-generator that breaks down the shape refinement process into two sub-goals, resulting in a disentangled but improved generation ability. We demonstrate the superiority of PointSea by conducting experiments on point cloud completion benchmarks with various data types. 

\noindent \textbf{Failure cases. }
Although PointSea effectively reconstructs detailed geometric features, transferring this ability to some complex out-of-distribution real-world scans is indeed challenging due to the domain gap. 
Fig.~\ref{fig:cases} shows two representative types of failure cases. The first row depicts a plant with disorganized leaves, and the second row shows a bed with an untidy quilt. Both examples exhibit complex topologies that are rarely observed in the training dataset. 
As a result, PointSea tends to rely on commonly observed patterns from the training data to reconstruct the point cloud, failing to capture the desired local details. Additionally, in the third row, the table contains outlier points from other objects in the scene, causing PointSea to generate noisy points in these extraneous structures.

\noindent \textbf{Potential solutions and future work.}
Considering the emerging network architectures or large-scale 3D generation or reconstruction model, we discussed the following three potential technical solutions to the aforementioned issue:

i) A promising approach is to train a large completion model with zero-shot generalization capabilities. Although PointSea leverages the scalable Transformer architecture as its main component, our early experiments indicate that simply stacking more attention layers does not necessarily improve completion performance. This observation is consistent with the scaling laws for large language models~\citep{kaplan2020scaling}. To enhance model robustness, generating additional high-quality training data using recent large-scale 3D datasets~\citep{objaverse, objaverseXL, yu2023mvimgnet} is a promising direction. These datasets have already played a pivotal role in advancing 3D generative models and could similarly benefit large point cloud completion models.

ii) Another solution is to utilize rich prior knowledge from existing large generative models.  As shown in our preliminary experiments (see Tab.~\ref{tab:ablation2Dbackbone}), integrating SVFNet with vision foundation models~\citep{lama,latentdiffusion,controlnet} may enhance the performance of PointSea.
In addition to 2D foundation models, 3D-aware generative models — including image-to-3D generation models, multi-view diffusion models, and native 3D diffusion models — can be employed to enhance shape completion capabilities.
Image-to-3D generation models typically produce high-quality local details and smooth surfaces. However, relying solely on single-view information may result in incomplete or less accurate outputs (see Fig.~\ref{fig:gen_comp}).
On the other hand, multi-view diffusion models provide richer, view-consistent geometric cues. For example, the recent method PCDreamer~\citep{wei2024pcdreamer} integrates this type of prior within a cross-modal completion framework to improve results.
Finally, inspired by PCDreamer, we believe that native 3D diffusion models (such as Clay~\citep{zhang2024clay}, 3DTopiaXL~\citep{chen2024primx}, and GaussianAnything~\citep{lan2024ga}) hold promising potential as a more direct 3D prior for point cloud completion in future research.

iii) In the current implementation of SDG, self-similarity is only considered within a single instance. However, similar patterns of missing regions, such as complex leaves, may occur across different shapes. Exploring cross-instance and cross-scene similarity offers an exciting research direction that could further enhance PointSea's ability to generalize across diverse scenarios.

\label{secCon}

\section*{Acknowledgements}
This work was supported by the National Natural Science Foundation of China (No. T2322012, No. 62172218, No. 62032011), the Shenzhen Science and Technology Program (No. JCYJ20220818103401003, No. JCYJ20220530172403007), and the Guangdong Basic and Applied Basic Research Foundation (No. 2022A1515010170).

\noindent\textbf{Data Availability}
All synthetic datasets can be accessed at \emph{\textcolor{magenta}{https://github.com/yuxumin/PoinTr/blob/master/DATASET.md}}. All real-world data can be accessed at \emph{\textcolor{magenta}{https://github.com/xuelin-chen/pcl2pcl-gan-pub}}. The scene datasets are publicly available at \emph{\textcolor{magenta}{https://github.com/JinfengX/CasFusionNet}}.

{\small
    \bibliographystyle{spbasic}
    \bibliography{pcd}
}

\end{document}